\documentclass[autheryear, times, 10pt]{elsarticle}

\usepackage{amsmath}
\usepackage[linesnumbered,ruled]{algorithm2e}
\usepackage{color}
\usepackage{diagbox}
\usepackage{float}
\usepackage{geometry}
\usepackage{booktabs}
\usepackage{graphicx}
\usepackage{hyperref}
\usepackage{lineno}
\usepackage{multirow}
\usepackage{makecell}
\usepackage{slashbox}
\usepackage{soul}
\usepackage{subfigure}
\usepackage{amsthm,amsmath,amssymb}
\usepackage{mathrsfs}
\usepackage[utf8]{inputenc}
\usepackage[english]{babel}
\usepackage[short]{optidef}
\usepackage{xcolor}
\usepackage{natbib}
\usepackage{nicematrix}
\usepackage{cases}
\usepackage[title]{appendix}
\usepackage{algpseudocode}
\usepackage{enumerate}
\usepackage[para,online,flushleft]{threeparttable}
\usepackage{titlesec}
\titleformat{\paragraph}[block]{\normalfont\itshape}{\theparagraph}{1em}{}

\usepackage{url}
\usepackage{verbatim}
\usepackage{hyperref}

\restylefloat{figure}

\newcommand{\redrevise}[1]{{\color{red}#1}}

\usepackage{xcolor}
\definecolor{bluerevise}{RGB}{76, 148, 216}  

\hypersetup{
    colorlinks,
    linkcolor={red!50!black},
    citecolor={blue!50!black},
    urlcolor={blue!80!black}
}

\biboptions{authoryear}
\DeclareMathOperator*{\argmin}{argmin}

\setcitestyle{notesep={; },round,aysep={},yysep={;}}
\journal{Transportation Research Part E: Logistics and Transportation Review}

\begin{document}
\begin{frontmatter}
\title{SPO-VCS: An End-to-End Smart Predict-then-Optimize Framework with Alternating Differentiation Method for Relocation Problems in Large-Scale Vehicle Crowd Sensing}

\author[address1]{Xinyu Wang}
\ead{xinyuu.wang@connect.polyu.hk}
\author[address1]{Yiyang Peng}
\ead{yiyang.peng@connect.polyu.hk}
\author[address1]{Wei Ma\corref{mycorrespondingauthor}}
\cortext[mycorrespondingauthor]{Corresponding author}
\ead{wei.w.ma@polyu.edu.hk}

\address[address1]{Civil and Environmental Engineering, The Hong Kong Polytechnic University \\ Hung Hom, Kowloon, Hong Kong SAR, China}

\begin{abstract}

Ubiquitous mobile devices have catalyzed the development of vehicle crowd sensing (VCS). In particular, vehicle sensing systems show great potential in the flexible acquisition of extensive spatio-temporal urban data through built-in smart sensors under diverse sensing scenarios. However, vehicle systems like taxis often exhibit biased coverage due to the heterogeneous nature of trip requests and varying routes. To achieve a high sensing coverage, a critical challenge lies in how to optimally relocate vehicles to minimize the divergence between the spatio-temporal distributions of vehicles and target sensing distributions. Conventional approaches typically employ a two-stage predict-then-optimize (PTO) process: first predicting real-time vehicle distributions and subsequently generating an optimal relocation strategy based on the predictions. However, this approach can lead to suboptimal decision-making due to the propagation of errors from upstream prediction. To this end, we develop an end-to-end Smart Predict-then-Optimize (SPO) framework by integrating optimization into prediction within the deep learning architecture, and the entire framework is trained by minimizing the task-specific matching divergence rather than the upstream prediction error. Methodologically, we formulate the vehicle relocation problem by quadratic programming (QP) and incorporate a novel unrolling approach based on the Alternating Direction Method of Multipliers (ADMM) within the SPO framework to compute gradients of the QP layer, facilitating backpropagation and gradient-based optimization for end-to-end learning. 
The effectiveness of the proposed framework is validated using two real-world taxi datasets ranging from mid-size to large-scale in Hong Kong and Chengdu, China. Utilizing the alternating differentiation method, the general SPO framework presents a novel concept of addressing decision-making problems with uncertainty, demonstrating significant potential for advancing applications in logistics and intelligent transportation systems.

\end{abstract}

  
\end{frontmatter}

\section{Introduction}
\label{sec: intro}

In recent years, the development of ubiquitous mobile devices such as mobile phones has significantly advanced the field of mobile crowd sensing (MCS) \citep{ganti2011mobile}. 
Compared with the traditional fixed sensor networks, MCS offers more flexible and extensive coverage of sensing information while reducing the installation and collection cost \citep{li2018high}. Individuals participating in MCS utilize the built-in smart sensors in mobile devices such as cameras and GPS, to sense the surroundings and gather city-wide information \citep{li2022unified}. 

The MCS can be categorized into community sensing (CS) and vehicle crowd sensing (VCS) based on different types of sensor hosts \citep{ji2023survey}. CS utilizes sensors installed in mobile devices, while VCS leverages the sensing capabilities of vehicles to collect city-wide data \citep{xu2019ilocus}. Host vehicles of VCS include taxis \citep{honicky2008n}, trams \citep{saukh2014route}, buses \citep{kang2016public}, and unmanned aerial vehicles (UAVs) \citep{li2022investigating}. 
In general, the advantages of VCS in terms of temporal duration, geographical coverage, reliability, and consistency make it particularly suitable for scenarios requiring continuous, full coverage, and high-quality data \citep{ji2023survey}.

Consequently, VCS is instrumental in domains that require high-precision data, such as infrastructure monitoring \citep{song2020deep, hasenfratz2015deriving, liu2022ubiquitous}, traffic state estimation \citep{guo2022sensing, zhu2014mobile} and traffic management \citep{ji2023trip, dai2023exploring}. 

Specifically, the \emph{vehicles} in the VCS framework are classified into two types: dedicated vehicles (DVs) and non-dedicated vehicles (NDVs). DVs, such as vehicles equipped with air pollution sensors and delivery drones with cameras, are fully guided by the dispatching center to prioritize sensing tasks. In contrast, NDVs, such as e-hailing vehicles and taxis, are under the complete control of their drivers and do not follow the guidance of the dispatching center. 
Upon receiving a sensing task, DVs will proceed immediately to the specific sensing locations. In contrast, NDVs operate under the control of drivers and are confined to their preferred cruising areas and routes. Moreover, NDVs can be converted into DVs through incentives such as monetary compensation.

\begin{figure}[H]
    \centering    \includegraphics[width=0.65\textwidth]{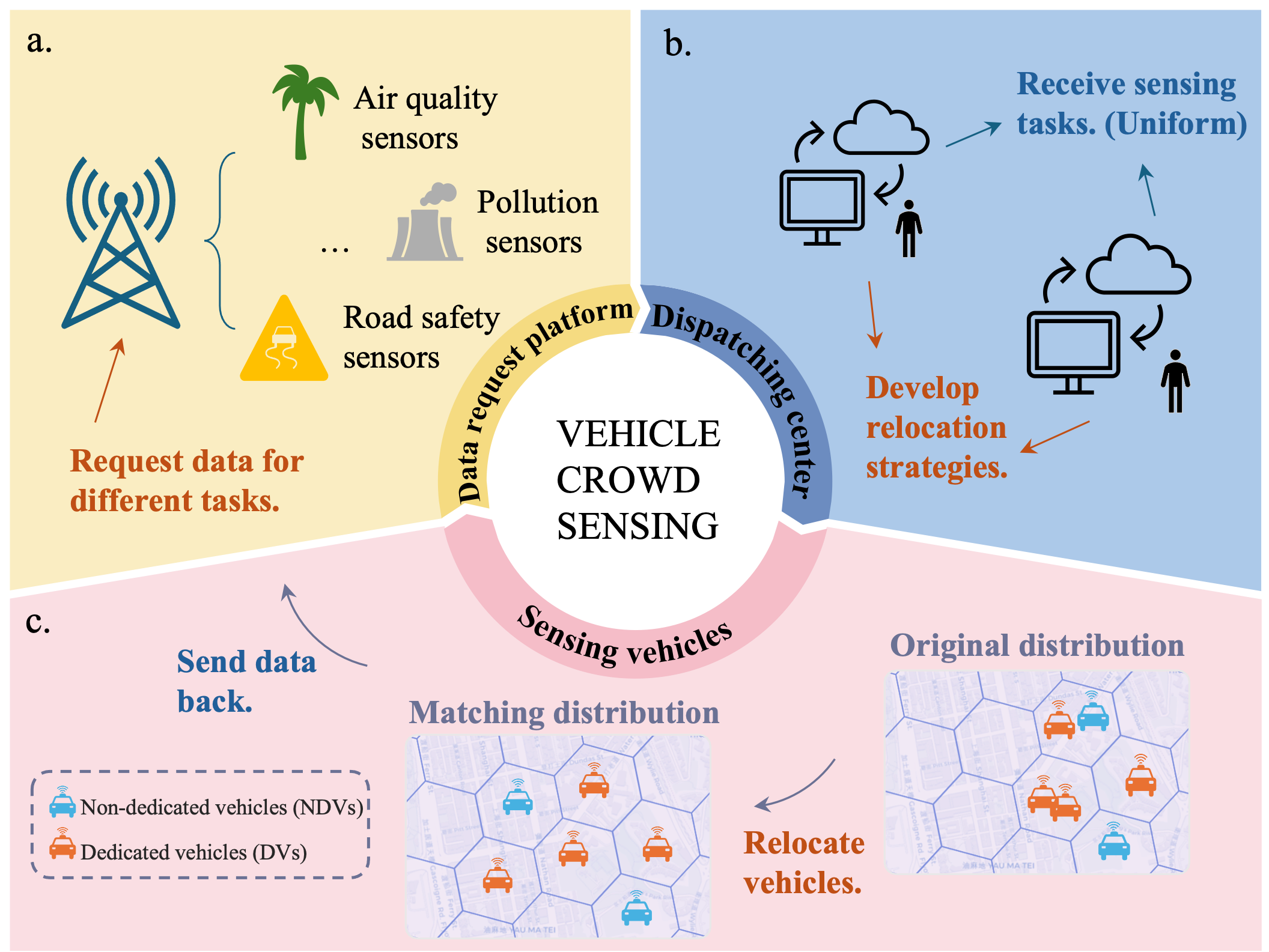}
    \caption{An illustration of the sensing procedure in the VCS framework.}
    \label{fig: overview}
\end{figure}

Figure \ref{fig: overview} provides an overview of the sensing procedure in the VCS framework. The \emph{data request platform} initially publishes sensing tasks. Upon receiving these tasks, the \emph{dispatching center} develops relocation strategies for DVs. During the subsequent time intervals, all the \emph{sensing vehicles}, including DVs and NDVs, move to the specified destinations to perform the assigned sensing tasks. After collecting the necessary data, the {sensing vehicles} will transmit sensing information back to the \emph{data request platform}.

In the domain of VCS, the target distribution of sensing tasks differs markedly from the spatio-temporal distribution of sensors. Target distributions typically vary based on specific sensing objectives and adapt over time and space. For instance, monitoring air quality may necessitate uniformly distributed data across a city \citep{bales2012citisense, chen2018pga}, whereas tasks focusing on industrial pollution may require the spatial distribution in Gaussian \citep{khedo2010wireless, paulos2007sensing}. In contrast, the spatio-temporal distributions of sensors are heavily influenced by the movements of host sensing vehicles \citep{anjomshoaa2018city}. 

Therefore, a critical challenge in VCS lies in the effective relocation of vehicles to align with the target sensing distribution. To be specific, the dispatching center needs to develop relocation strategies for DVs to minimize the divergence between the target sensing distributions and vehicle distributions.
The state-of-the-art approach to addressing the vehicle relocation problem in VCS is a two-stage predict-then-optimize (PTO) method, where a prediction model is first established to forecast the movements of NDVs \citep{chang2023smart, simchi2014om, mivsic2020data}. The prediction is then used as the input for the downstream optimization problem related to DVs. 
However, a key limitation of the PTO approach is that training the prediction model is mainly based on empirical error minimization, and metrics such as Mean Squared Error (MSE) and Mean Absolute Error (MAE) are used. 
The trained prediction model may not necessarily yield optimal strategies in the subsequent optimization stage, because the focus of the optimization stage is task-specific, which may not align well with the metrics of MSE or MAE \citep{bengio1997using, ford2015beware, elmachtoub2022smart, yan2022integrating}. This discrepancy highlights the need for more integrated approaches that bridge prediction and optimization phases coherently in the context of VCS.

Recently, the end-to-end Smart Predict-then-Optimize (SPO) framework has been proposed and attracted wide attention \citep{elmachtoub2022smart}. This framework integrates prediction and optimization within a unified deep-learning architecture. Specifically, the optimization module operates as a differentiation layer, mapping inputs directly to optimal solutions without requiring explicit closed-form solutions. This approach enables the entire neural network to be trained by directly minimizing task-specific errors. However, the application of the SPO framework to address challenges such as vehicle relocation has not been extensively explored.

The main challenge in applying the SPO framework is integrating the optimization layer seamlessly within the deep learning architecture.
A widely used approach is the implicit differentiation, where the Karush-Kuhn-Tucker (KKT) conditions of the optimization problems are differentiated to obtain first-order derivatives of the optimal solution with respect to the parameters. However, this method can be computationally demanding in directly computing the Jacobian matrix of the KKT conditions for large-scale problems \citep{diamond2016cvxpy, agrawal2019differentiable, tang2022pyepo}.

In this paper, we advocate an alternative approach - explicit unrolling, which avoid the direct computation of the Jacobian matrix in the KKT conditions and compute the corresponding gradients iteratively, enhancing computational efficiency and suitability for large-scale problems \citep{sun2023alternating, li2020end}. Nevertheless, the majority of unrolling approaches are predominantly applicable to unconstrained problems. In the task of VCS, relocation challenges are usually formulated as constrained optimization problems. To this end, we incorporate the Alternating Direction Method of Multipliers (ADMM) to solve constrained optimization problems by decomposing the original problem into a series of unconstrained sub-problems \citep{boyd2011distributed}. Combining the ADMM with differentiation methods and unrolling approaches can potentially address the challenge of integrating large-scale constrained convex optimization problems into deep learning architectures. 

To conclude, there are two major issues with the predict-and-optimize problem in VCS. First, the conventional PTO framework for vehicle relocation problems may not generate an optimal relocation strategy, resulting in a substantial divergence between the desired target distribution and the actual vehicle distribution. Second, embedding an optimization layer within neural networks is usually computationally intensive and impractical for large-scale networks, and existing unrolling approaches are often inadequate for handling constrained optimization problems effectively.

To address these challenges, we develop an innovative SPO framework coupled with an alternating differentiation method for vehicle relocation problems in VCS, aiming to achieve optimal sensing coverage. Specifically, we formulate the vehicle relocation problem based on constrained quadratic programming (QP) and embed the QP layer into neural networks. To our knowledge, this is the first time that the SPO framework coupled with an alternating differentiation method has been applied in vehicle relocation in mobile sensing. Our contributions are outlined as follows:

\begin{itemize} 
    \item \emph{A novel end-to-end Smart Predict-then-Optimize framework for vehicle relocation problems}. We first time formulate the vehicle relocation problem using an SPO framework in deep learning. Specifically, the vehicle relocation problem is formulated as a constrained QP, where the parameters in its objectives are predicted using deep learning.

    \item \emph{An alternating differentiation approach based on the ADMM for constrained optimization problems in large-scale networks}. A novel unrolling approach is proposed to obtain the gradients of the constrained QP iteratively for large-scale networks. Specifically, the forward pass is solved using the ADMM, and the backpropagation is achieved through alternating differentiation. 

     \item \emph{Two large-scale real-world experiments}. Numerical experiments on two real-world datasets across five network sizes from mid-size to large-scale are conducted. Superior matching performance and computational efficiency of the proposed SPO framework are demonstrated.

\end{itemize}

The remainder of the paper is organized as follows: Section \ref{sec: lr} reviews the related work on the vehicle relocation problem in mobile sensing. Section \ref{sec: model} introduces the SPO framework in vehicle relocation, including the prediction module, the optimization module, the integrated model, and an unrolling approach based on the alternating differentiation method for the SPO framework. Section \ref{sec: pseudo code} presents the overall solution algorithm of the end-to-end SPO framework. Section \ref{sec: experiment} evaluates the proposed framework in two real-world scenarios and compares the method with the baseline approaches. Section \ref{sec: conclusion} concludes the results and proposes future directions.

\section{Literature Review}
\label{sec: lr}

This section provides a comprehensive review of the literature related to the vehicle relocation problem in mobile sensing, the end-to-end SPO framework and its differentiation approach, and applications of the SPO framework in the field of transportation.

\subsection{Vehicle relocation in mobile sensing}

Pervasive crowd sensing plays an important role in intelligent transport systems, logistics and smart city applications \citep{ji2023survey, dai2023exploring, yang2025joint}. Given their relatively lower installation costs and extended operational duration, taxis are the most commonly utilized hosts in the VCS. However, due to heterogeneous trip patterns, the spatio-temporal distributions of these vehicles are often biased. To enhance sensing quality, numerous scholars have investigated intervention measures to effectively relocate vehicles through the development of incentivizing schemes \citep{fan2019joint, xu2019ilocus, chen2020pas} or proactive scheduling \citep{masutani2015sensing}.

Existing research on solving the conventional vehicle relocation problem can be categorized into three types: the steady-state method, the model-free method, and the real-time optimization method \citep{lei2020efficient, qian2022drop}. The steady-state method modeled the vehicle relocation system through a queuing theoretical framework and solved the optimal relocation strategy when the system reaches a steady state \citep{pavone2012robotic, sayarshad2017non, chuah2018optimal}. A significant limitation of this approach is the inadequate capacity to manage the complexities of large-scale problems.  In contrast, the model-free method, based on deep reinforcement learning (DRL), has emerged as a promising approach for its merits in addressing large-scale vehicle relocation challenges \citep{qian2022drop, jiao2021real, shou2020reward, qin2022reinforcement}. 

Another classical approach in vehicle sensing is the real-time optimization method, which formulates the vehicle relocation problem from an optimization perspective and identifies optimal strategies through various solutions \citep{kek2009decision, huang2018solving}. \citet{miao2016taxi} and \citet{zhang2016model} formulated the vehicle scheduling and routing problem as a mixed integer program (MIP) and utilized the model predictive control (MPC) method to solve it. Recent advancements have also incorporated deep learning techniques to forecast future travel patterns, employing a two-stage PTO framework for the vehicle relocation problem \citep{weikl2013relocation, chang2022cooperative, chang2023smart}. This method first estimates the spatio-temporal vehicle patterns using deep learning, followed by solving the downstream optimization problem with the prediction results as inputs. However, a notable limitation of the PTO approach is that training the prediction model solely based on prediction error can result in suboptimal decision-making, as opposed to directly minimizing the decision error \citep{agrawal2019differentiable, elmachtoub2022smart}. Even with optimal predictions yielding minimal prediction errors, the resulting optimization may not be optimal. Our study extends the real-time PTO framework by integrating optimization directly within the prediction process through a deep learning architecture.

Table \ref{tab: reference} summarizes related works of the vehicle relocation problem in mobile sensing and their fundamental settings. In a nutshell, our work differs from the previous research in three key aspects. First, unlike most existing literature that fully controls vehicles, we classify vehicles into dedicated and non-dedicated types, allowing the non-dedicated ones to operate freely. Second, to our knowledge, we are the first to apply the SPO framework to vehicle relocation in mobile sensing and validate it on large-scale networks. Third, we evaluate the framework across multiple target distributions, an aspect rarely explored in previous studies.

\begin{table}[h]
    \caption{The related work of vehicle relocation in mobile sensing.}
    \label{tab: reference}
    \centering
    \addtolength{\tabcolsep}{10pt}
    \begin{threeparttable}
        \begin{tabular}{ccccccc}
        \toprule
            References & \makecell[c]{Fleet\\type} & \makecell[c]{Control \\ ratio}  & \makecell[c]{Target \\ distribution} & \makecell[c]{Solution\\approach}  \\
        \midrule
         \citet{miao2016taxi} & Taxi & Fully & Reality  & MPC \\
         \citet{xu2019ilocus} & Taxi &  Partially  & U, G, GM  & MPC \\
         \citet{wang2019hytasker} & Not specific &  Partially & Random   & PTO  \\
         \citet{chen2020pas} & Taxi &  Partially & Balanced  & DRL \\
         \citet{fan2021towards} & Taxi, DV & Fully  & N,U  & OPT \\
          \citet{jiao2021real} & Ride-hailing & Fully & - & DRL \\
         \citet{xu2021task}  & UAV & Fully & R   &   OPT \\
          \citet{qian2022drop} & Taxi & Fully & U  & DRL \\
          \citet{chang2023smart} & Bike & Fully  & Predicted  & PTO \\
         \citet{jiang2023ship} & Taxi & Fully & U  &  OPT \\
         \textbf{Ours} & Taxi &  Partially  & U, G, GM & \textbf{SPO} \\
         \bottomrule
        \end{tabular}
    \begin{tablenotes}
        \item U: Uniform distribution; G: Gaussian distribution; GM: Gaussian mixture distribution; N: Normal distribution; OPT: optimization without prediction.
    \end{tablenotes}
    \end{threeparttable}
\end{table}

\subsection{The end-to-end SPO framework and its differentiation methods}

The SPO framework is an emerging paradigm in the data-driven optimization field that can leverage deep learning tools with potential applications in inventory management and electric grid scheduling etc. \citep{elmachtoub2022smart, donti2017task}. Unlike traditional neural networks that utilize explicit formulations, the SPO framework integrates an optimization module as an implicit differentiation layer. This approach maps the inputs of the implicit functions within the optimization layer to optimal solutions. Consequently, training the optimization layer is not easy due to the absence of explicit closed-form derivatives.

To compute the gradients of the optimization layer, two principal methods are employed: the implicit differentiation method and the explicit unrolling approach. The implicit method involves calculating the first-order derivatives of the optimal solution concerning the parameters, typically by directly differentiating the KKT conditions \citep{gengattention, geng2021training}. Tools such as CVXPY \citep{diamond2016cvxpy}, Cvxpylayer \citep{agrawal2019differentiable}, PyEPO \citep{tang2022pyepo}, and OptNet \citep{amos2017optnet} facilitate this process for various problem types, including linear programming (LP), integer programming (IP) \citep{ferber2020mipaal}, and QP. However, implicit differentiation methods necessitate extensive computation of the Jacobian matrix, rendering them unsuitable for large-scale network problems \citep{sun2023alternating}. An alternative approach is the alternating approach, which applies an iterative first-order gradient method. Existing research on alternating approaches is limited.  A notable contribution is by \citet{sun2023alternating}, which proposed an alternating differentiation method based on the ADMM to differentiate convex optimization problems with polyhedral constraints efficiently and recursively. The alternating differentiation method decouples the optimization layers in an alternative manner, thereby reducing the dimension of the Jacobian Matrix. \citet{li2020end} proposed both implicit and explicit unrolling methods for differentiating a Variational Inequalities (VIs) layer with a fixed point method. The implicit method leverages the sensitivity of the solutions to the VIs, while the explicit approach unrolls the projection method to solve the VIs.

Prior research has explored various approaches to enable differentiable optimization within the SPO framework for integer programming problems, which are broadly categorized into three main types, including linear programming relaxations and smoothing \citep{zhang2024end, mandi2020smart}, incorporating the revised algorithms such as cutting-plane \citep{ferber2020mipaal}, and utilizing surrogate loss functions such as Perturbed Fenchel-Young Loss (PFYL) and SPO+ loss \citep{berthet2002learning,  elmachtoub2022smart}. These approaches collectively provide effective differentiable substitutes that facilitate end-to-end learning for integer optimization problems and advance the integration of integer programming into the SPO framework. 

In summary, implicit differentiation methods directly obtain derivatives but entail significant computational costs for large-scale problems, making them more suitable for simpler optimization problems, such as LP with few constraints or relaxed IP \citep{mandi2020smart, elmachtoub2022smart}. Conversely, explicit differentiation methods unroll the differentiation procedure, reducing the dimension of the inverse Jacobian matrix, and can be more suitable for large-scale and complex convex optimization problems, such as QP.

\subsection{Applications of the SPO framework}
Recent studies have increasingly investigated the application of the end-to-end SPO framework to traffic operations and management \citep{yan2021emerging, lin2024unlocking, yang2025dff, yang2025decision}, in the fields of maritime transportation \citep{yan2020semi, tian2023smart, yan2023extended, yang2024efficient} and logistics \citep{liu2021time, qi2023practical}. \citet{yan2020semi} developed a semi-SPO framework for efficient ship inspections by utilizing a surrogate mean squared difference in overestimating the number of deficiencies, rather than minimizing the mean squared error. This approach involved first predicting deficiency numbers for each inspector per ship and then constructing an integer optimization model to allocate inspectors to deficient ships. It is noteworthy that in their semi-SPO framework, prediction and optimization remain distinct processes. Similarly, \citet{yang2024efficient} follows the SPO criterion and proposes a task-specific metric named cumulative detected deficiency number (CDDN) to evaluate the efficiency and effectiveness of port state control inspection. However, the above studies focus on proposing novel SPO metrics rather than embedding an optimization layer in the neural network. 

Although the SPO framework has been applied to various fields \citep{yan2020semi, liu2021time, yan2021emerging, tian2023smart, yan2023extended, qi2023practical, yang2024efficient, zhang2024end}, its applications in vehicle relocation are overlooked. A closely related study of embedding optimization layer to ours is by \citet{zhang2024end}, which applied the end-to-end SPO framework to express pickup and delivery systems. They constructed a Graph Convolutional Network (GCN)-based model to predict order quantities and formulated a K-means clustering model to determine the optimal Areas of Interest (AOI) assignment for couriers. They utilized implicit differentiation by directly differentiating the KKT conditions to enable backpropagation within the neural network.

\begin{table}[H]
    \caption{Comparison of the related work in the SPO framework in transportation within the deep learning architecture.}
    \label{tab: spo comparison}
    \centering
    \small
    \begin{tabular}{cccccc}
        \toprule
        References & \makecell[c]{Optimization\\layer} & Constraints & \makecell[c]{Differentiation\\method} & Dimension & Scalability \\
        \midrule
        \citet{zhang2024end} & LP & $\surd$ & Implicit by KKT & 35, 100 & Small \\ 
        \textbf{Ours} & QP & $\surd$ & \makecell[c]{Alternating differentiation} &  \makecell[c]{$2,025\ (45 \times 45),$\\ $4,624\ (68 \times 68),$ \\$3,600 \ (60 \times 60)$\\$6,400\  (80 \times 80)$\\$10,000 \ (100 \times 100)$} & Large \\
        \bottomrule 
    \end{tabular}
\end{table}

Table \ref{tab: spo comparison} presents a comparison of studies employing the SPO framework in transportation within the deep learning architecture. Notably, compared to the study by \citet{zhang2024end}, our work demonstrates increased complexity in the formulation of the QP layer, characterized by a higher dimensionality of decision variables, and distinctively employs the alternating differentiation method to address large-scale problems.

\section{The SPO framework} 
\label{sec: model}

In this section, we first present a comprehensive overview of the SPO framework, detailing its core components: (1) the prediction module for demand estimation, (2) the optimization module for decision-making, and (3) the backpropagation mechanism that enables end-to-end learning. Second, we demonstrate the application of the proposed SPO framework to vehicle relocation problems in VCS, providing complete mathematical formulations for each sub-module and the integrated SPO framework. Finally, we develop the alternating differentiation method and derive specialized solution schemes to efficiently solve the relocation problem while maintaining gradient flow for backward propagation.

\subsection{Overview of the SPO framework}

\begin{figure}[h]
    \centering
    \includegraphics[width=0.9\textwidth]{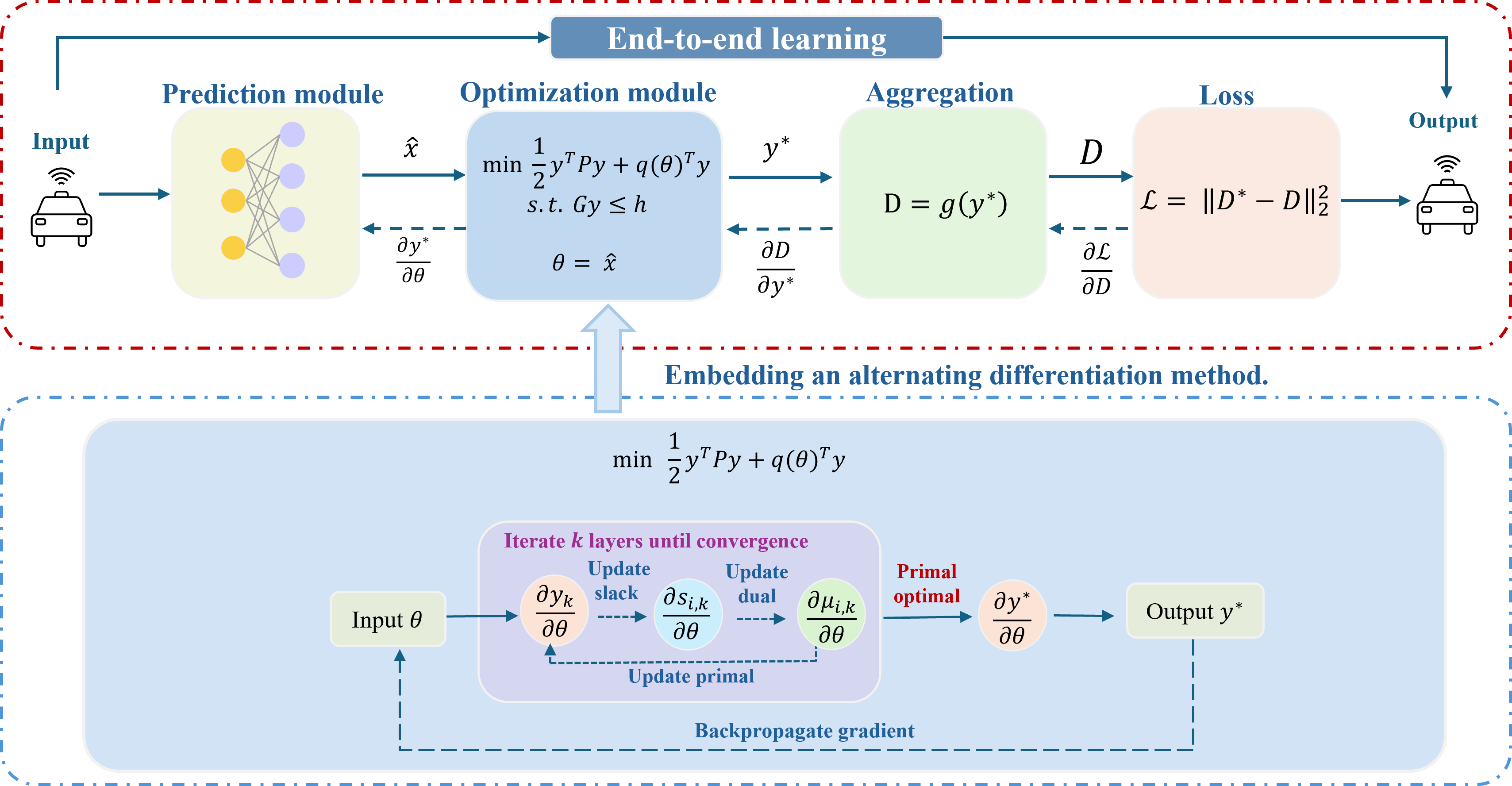}
    \caption{The end-to-end SPO framework of the vehicle sensing problem.}
    \label{fig: end to end}
\end{figure}

Figure \ref{fig: end to end} presents the end-to-end SPO framework, which consists of two essential modules: a prediction module and an optimization module. The prediction module uses a machine learning model to generate parameter estimates, denoted as $\hat{x}$, which configure the optimization module by serving as input parameters $\theta$. The optimization module, which is formulated as a quadratic program with objective function $\frac{1}{2}y^TPy  + q(\theta)^Ty$ and linear constraints $Gx\leq h$, yields the optimal solution $y^*$. Note that $x, \theta,y, P, q\in \mathbb{R}^n$, $G,h \in \mathbb{R}^m$. Following optimization, an optional aggregation module performs post-processing computations on $y$ (e.g., summation or averaging) to produce the final decision variable $D$, which is evaluated against the target $D^*$ through a loss function $\mathcal{L}$. The SPO framework aims to minimize the discrepancy between decision $D$ and target $D^*$, ensuring alignment with decision-making objectives.

Specifically, the optimization module incorporates an alternating differentiation algorithm designed to facilitate gradient flow through the module. The algorithm employs the ADMM to compute the optimal solution $y^*$ while simultaneously updating slack variables $s$ and dual variables $\mu$. The alternating differentiation procedure iterates in a total of $K$ layers, with each layer executing alternating updates until the solution converges to the optimal value. During the backward pass, gradients propagate through the optimal values involving primal variables $\mathbf{y}$, slack variables $s$ and dual variables $\mu$, to update the model parameters $\theta$. This complete gradient flow enables end-to-end learning of the entire SPO framework with alternating differentiation methods.

We then introduce the application of the SPO framework to the vehicle relocation problems in crowd sensing. As previously discussed, DVs in the VCS system adhere to designated routes under the direction of the dispatching center, whereas NDVs independently select their cruising areas and pathways. Consequently, when the dispatching center receives sensing requests from the upstream platform, it lacks prior knowledge regarding the future spatial distribution of NDVs, despite their considerable impact on the overall vehicle distribution. To devise an effective relocation strategy for DVs, it is crucial to predict the distribution of NDVs in advance. This predicted demand is then incorporated into the optimization module as a parameter.

\subsection{Vehicle demand prediction model for NDVs}

This section first introduces the spatio-temporal neural network, then presents the design of the prediction model for NDVs in the SPO framework.

\subsubsection{Spatio-temporal vehicle movement prediction}

The vehicle demand varies dynamically over time and space. A common approach to manage this variability is discretizing the study area and time length into small intervals for easy implementation of specialized spatio-temporal deep learning models \citep{ke2021joint}.

In this paper, the vehicle prediction problem is modeled on a spatio-temporal scale. For the spatial dimension, the study area is partitioned into various regular hexagon grids based on geographical information \citep{ke2018hexagon}. List of notations and dimensions for variables are provided in \ref{sec: appendix A} (See Table \ref{apd: list of notations} and Table \ref{apd: variable vectorization}). For the temporal dimension, each day is uniformly divided into equal time intervals. Consider a spatial network $\mathcal{G = (V, E)}$, where $\mathcal{V}$ denotes the set of vertices consisting of the set of origin grids $I$ and the set of destination grids $J$, i.e., $\mathcal{V = I \cup J}$, such that $\vert I \vert = \vert J \vert = N$. The set of edges $\mathcal{E}$  edges connects all adjacent grids. 

Let $V$ denote the vehicle class, with each class $v$ defined as $v = { c, f, a }$, where $c$ represents controllable DVs, $f$ represents free NDVs, and $a$ encompasses all vehicles. The set of time intervals is denoted as $T$, with each interval of length $\delta$ for each time interval $\tau \in T$. For each time interval $\tau$, the spatial distribution of each vehicle type $v$ is defined as $\mathbf{D}_{v}^{\tau} \in \mathbb{R}^{N}$. To be precise, $\mathbf{D}_{v}^{\tau} = 
[D_{v,1}^{\tau},...,D_{v,i}^{\tau};...,D_{v,I}^{\tau}]^{\prime} \in \mathbb{R}^{N}$, where $D_{v,i}^{\tau}$ represents the vehicle demand of vehicle type $v$ in grid $i$ at time $\tau$, and $^{\prime}$ is the transpose of a matrix. Then the matching distribution of all vehicles at time $\tau$ is represented by $\mathbf{D}_a^{\tau} = \mathbf{D}_{f}^{\tau} + \mathbf{D}_{c}^{\tau}$.

Based on the above notations, the short-term demand prediction model of NDVs is defined as follows: Given the historical observations of NDVs over the past $m$ time intervals $[D_{f}^{\tau-m+1}, \ldots, D_{f}^{\tau}]$, the goal is to predict the spatial demand $\mathbf{D}_{f}^{\tau+1}$ for the next time interval $\tau+1$.

\subsubsection{TGCN-based prediction model for NDVs}

The Temporal Graph Convolutional Network (TGCN) \citep{zhao2019t}, which integrates the Graph Convolutional Network (GCN) and the Gated Recurrent Unit (GRU), is capable of concurrently capturing the inherent spatial dependencies of vehicle demand in adjacent regions and the temporal dynamics. Accordingly, we develop a two-layer prediction model based on the TGCN architecture to forecast the real-time demand of NDVs. The model incorporates a Rectified Linear Unit (ReLU) activation function and is defined as follows:

\begin{equation}
\label{eq: tgcn model}
    \hat{D}_{f,i}^{\tau+1} = \text{ReLU}(\text{TGCN}(\text{ReLU}(\text{TGCN}(D_{f,i}^{\leq\tau}, A_{f,i}^{\leq\tau}, \hat{H}_{f,i}^{\leq\tau})))),
\end{equation}
where the $\hat{D}_{f,i}^{\tau+1}$ is the predicted demand of NDVs at time $\tau + 1$, which is also the output of the prediction module, $A_{f,i}^{\leq\tau}$ is the adjacency information, $\hat{H}_{f,i}^{\tau}$ is the hidden information. The hat symbol, $\hat{\cdot}$, indicates the variable is an estimator for the true(unknown) variable. $\text{ReLU}(\cdot) = max(0,\cdot)$.

We define the prediction loss in the prediction module in Eq. (\ref{eq: prediction loss}):

\begin{equation}
\label{eq: prediction loss}
    \mathscr{L}_1 = \lVert {\mathbf{D}_{f}^{\tau+1} - \hat{\mathbf{D}}_{f}^{\tau+1}} \rVert _2 ^2,
\end{equation}
where $\mathbf{D}_{f}^{\tau+1}$ is the actual distribution of the NDVs at time $\tau+1$.

Note that any spatio-temporal prediction model can be used in the SPO framework, here we employ the classical TGCN model as an example.

\subsection{Vehicle relocation model for DVs}
This section initially discusses the variable relationship between prediction and optimization modules, then presents the formulation of the vehicle relocation problem and its vectorized representation.

\subsubsection{Connection between prediction and optimization variables}
\begin{figure}[h]
    \centering
    \includegraphics[width=0.65\textwidth]{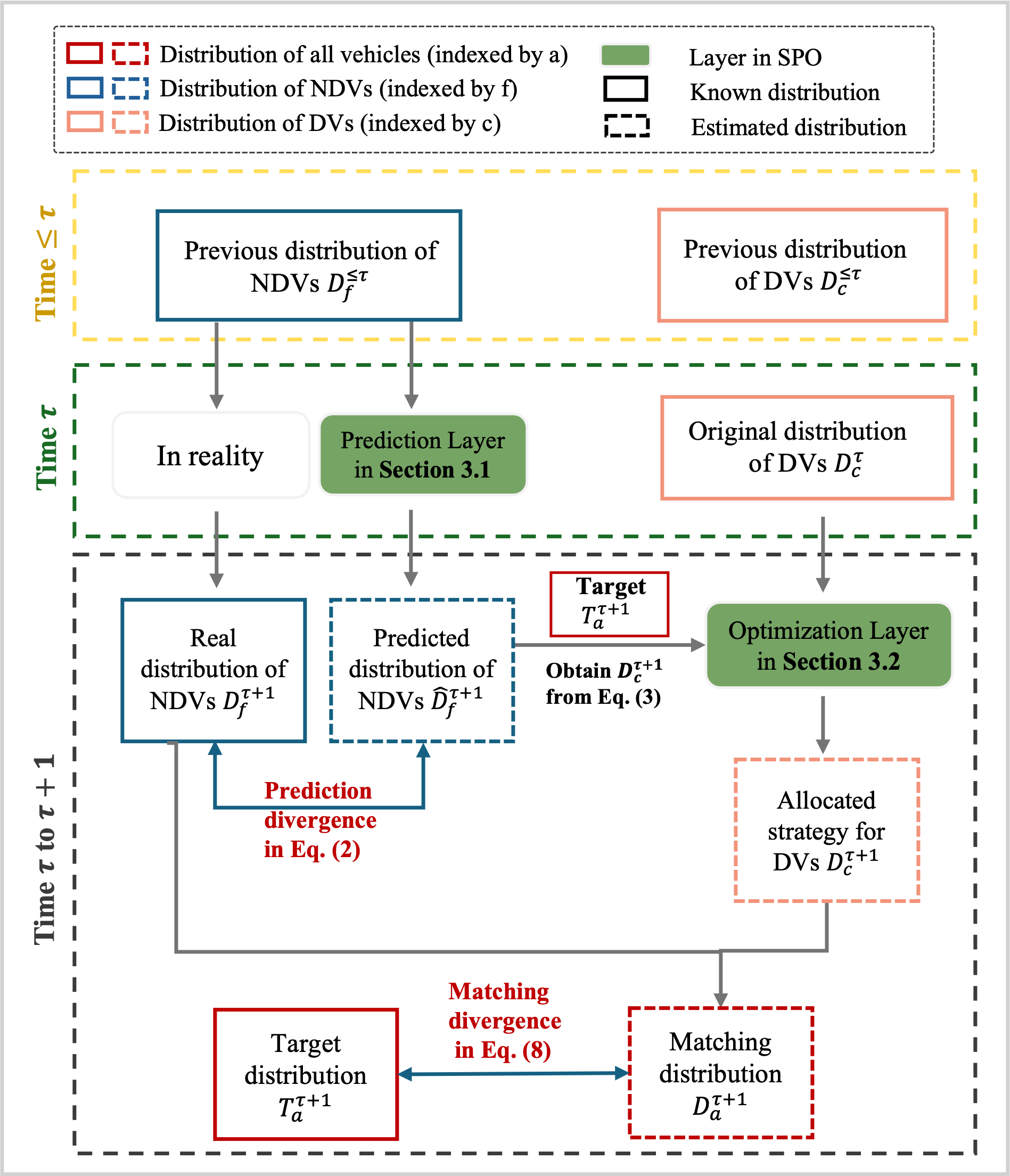}
    \caption{The variable relationship in different modules.}
    \label{fig: variable relationship}
\end{figure}

Before proposing the optimization model, we first illustrate the connection among the various variables between prediction and optimization modules, as shown in Figure \ref{fig: variable relationship}. The output of the prediction layer $\hat{\mathbf{D}}_{f}^{\tau+1}$ is input in the optimization layer as a known parameter.

After predicting the movements of NDVs, we can easily obtain the required (estimated) distribution for DVs by a simple subtraction of the given target distribution $\mathbf{T}_a^{\tau+1}$ and the predicted distribution $\hat{\mathbf{D}}_f^{\tau+1}$ in Eq. (\ref{eq: sum for matching}):

\begin{equation}
\label{eq: sum for matching}
    \hat{\mathbf{D}}_c^{\tau+1} = \mathbf{T}_a^{\tau+1} - \hat{\mathbf{D}}_f^{\tau+1},  
\end{equation}
where $\mathbf{T}_a^{\tau+1}$ is the target distribution for all vehicles at time $\tau+1$.

\subsubsection{Formulation of the vehicle relocation model}

This section presents the QP formulation of the relocation model for DVs within the optimization module. We regard the original distribution of DVs $\mathbf{D}_c^{\tau}$ as \textit{supply} and the required distribution of DVs $\mathbf{\hat{D}}_c^{\tau+1}$ as \textit{demand} for simplicity. The goal in the subsequent optimization module is to develop an optimal vehicle relocation strategy that satisfies the future \textit{demand} with the current \textit{supply} by considering the total incentive cost  budget, accessibility constraints, and supply limitations. 

Denote $x_{c,ij}^{\tau} \in \mathbb{R}^{N \times N} $ as the dedicated vehicle flow from grid $i$ to grid $j$ during $\tau$ to $\tau +1$. Then the vehicle relocation problem is formulated in Eq. (\ref{eq: original qp}):

\begin{mini!}{x_{c,ij}^{\tau}}
    {Z_0 = \frac{1}{2} \sum_j \lVert {\sum_i x_{c,ij}^{\tau} - \hat{\mathbf{D}}_{c,j}^{\tau+1}} \rVert _2^2,}
    {\label{eq: original qp}}{\label{obj: original form}}
    \addConstraint{ \sum_j x_{c,ij}^{\tau}}{\leq \mathbf{D}_{c,i}^{\tau}}, \forall i\in I, \tau \in T {\label{cst: supply}}
    \addConstraint{(m_{c,ij}^{\tau} - \delta) x_{c,ij}^{\tau}}{\leq 0}, \forall i\in I,j\in J, \tau \in T {\label{cst: time}}
    \addConstraint{\sum_i \sum_j w_{c,ij}^{\tau} x_{c,ij}^{\tau}}{\leq R^{\tau}}, \forall \tau \in T {\label{cst: budget}}
    \addConstraint{x_{c,ij}^{\tau}}{\geq 0},  \forall i \in I,j \in J, \tau \in T, {\label{cst: non nag}}
\end{mini!}
where $m_{c,ij}^\tau \in \mathbb{R}^{N \times N}$ is the travel time from grid $i$ to $j$ at time $\tau$; $w_{c,ij}^\tau \in \mathbb{R}^{N \times N}$ is the incentive cost from grid $i$ to $j$ at time $\tau$; and $R$ is the maximum total incentive budget.  The objective in Eq. (\ref{obj: original form}) aims to satisfy the \emph{demand} in all regions. The constraints of the problem include supply constraint in Eq. (\ref{cst: supply}), time constraint in Eq. (\ref{cst: time}), budget constraint in Eq. (\ref{cst: budget}) and non-negative constraint in Eq. (\ref{cst: non nag}). The supply constraint  in Eq. (\ref{cst: supply}) ensures that the total number of DVs at time $\tau+1$ does not exceed the available supply. The time constraint in Eq. (\ref{cst: time}) checks whether the actual travel time exceeds the desired relocation time $\delta$. The budget constraint in Eq. (\ref{cst: budget}) ensures that the incentive cost of all DVs should not surpass the total budget $R$. 

\subsubsection{Vectorizing the relocation formulation}

To seamlessly integrate the optimization module into the neural network, all two-dimensional variables and parameters will be flattened and vectored into one dimension. Specifically, $x_{c,ij}^{\tau}\in \mathbb{R}^{N\times N}$, $w_{c,ij}^{\tau}\in \mathbb{R}^{N\times N}$ are converted to $\mathbf{y}^{\tau} \in \mathbb{R}^{N^2}$, $\mathbf{C}^{\tau}\in \mathbb{R}^{N^2}$, respectively. For simplicity, we omit the superscript $\tau$, therefore $\mathbf{y}^{\tau} \in \mathbb{R}^{N^2}$, $\mathbf{C}^{\tau}\in \mathbb{R}^{N^2}$ is replaced by $\mathbf{y}$, $\mathbf{C}$, as is shown below: 

\begin{align}
  \mathbf{y} & =\left( y_1,\  y_2,\  ...,\  y_N\ |\  y_{N+1},\ \ y_{N+2},\  ...y_{2N}|\ y_{N^2-N+1},\ y_{N^2-N+2},\ ...,\ y_{N^2} \right)^T \\
  & = \left( x^{\tau}_{11},\  x^{\tau}_{12},\ ...,\ x^{\tau}_{1N}|\ x^{\tau}_{21},\  x^{\tau}_{22},\ ...,\ x^{\tau}_{2N}|\ x^{\tau}_{N1} ,\ x^{\tau}_{N2} ,\ ...,\  x^{\tau}_{NN} \right)^T \ \in \mathbb{R}^{N^2}, \\
  \mathbf{C} &= \left( w^{\tau}_{11},\ w^{\tau}_{12},\ ...,\  w^{\tau}_{1N} |\ 
w^{\tau}_{21},\ w^{\tau}_{22},\ ...w^{\tau}_{2N}|\ w^{\tau}_{N1},\ w^{\tau}_{N2},\ ...,\ w^{\tau}_{NN} \right)^T .
\end{align}

The vehicle relocation problem in Eq. (\ref{eq: original qp}) is then reformulated into a general standardized quadratic form, as presented in Eq. (\ref{eq: y}).

\begin{mini}{\mathbf{y}}
    {Z_1 = \frac{1}{2} \mathbf{y}^{\prime} \mathbf{P} \mathbf{y} + \mathbf{q}^{\prime} \mathbf{y},}{\label{eq: y}}{}
    \addConstraint{\mathbf{G_1}\mathbf{y}}{\leq \mathbf{h_1}}
    \addConstraint{\mathbf{G_2}\mathbf{y}}{\leq \mathbf{h_2}}
    \addConstraint{\mathbf{G_3}\mathbf{y}}{\leq \mathbf{h_3}}
    \addConstraint{\mathbf{G_4}\mathbf{y}}{\leq \mathbf{h_4},}
\end{mini}
where $Z_1 = Z_0 - \frac{{{}\hat{\mathbf{D}}_c^{\tau+1}}^{\prime}
\hat{\mathbf{D}}_c^{\tau+1}}{2}$, $\mathbf{P} = \mathbf{A}^{\prime}\mathbf{A}$, $ \mathbf{q}= -\mathbf{A}^{\prime} \hat{\mathbf{D}}_c^{\tau+1}$, $\mathbf{G_1} = \mathbf{B}$, $ \mathbf{G_2} = \text{diag}(m^\tau_{11} - \delta, m^\tau_{12} -\delta, ..., m^\tau_{NN} - \delta) \in \mathbb{R}^{N^2}$, \\ $\mathbf{G_3} = \mathbf{C}^{\prime}$, $ \mathbf{G_4} = -\boldsymbol{I_{N^2}}$,  $\mathbf{h_1} = \mathbf{{D}_{c}^{\tau}}$,  $\mathbf{h_2} = \boldsymbol{0}$,  $\mathbf{h_3} = \mathbf{R}$,  $\mathbf{h_4} = \boldsymbol{0}$. $\text{diag}(\cdot)$ is the diagonal matrix, $I_{N^2}$ is the Identity matrix in dimension $N^2$. $\mathbf{A}$ and $ \mathbf{B}$ are sparse matrices, and, \\
$ \mathbf{A} = \left(
\begin{array}{ccc|ccc|c|ccc}
    1 & 0 \cdots & 0 & 1& 0 \cdots &0 &  & 1 & 0 & \cdots 0\\
    0 & 1 \cdots & 0 & 0& 1 \cdots &0 & & 0 & 1 & \cdots 0\\
    \vdots & \vdots & \vdots & \vdots & \vdots & \vdots & \cdots & \vdots & \vdots & \vdots \\ 
    0 & 0 \cdots & 1 & 0& 0 \cdots &1 & & 0 & 0 & \cdots 1\\
\end{array}
\right)_ {N \times N^2}
$,
$ \mathbf{B} = \left(
\begin{array}{ccc|ccc|c|ccc}
    1 & 1 \cdots & 1 & 0& 0 \cdots &0 &  & 0 & 0 & \cdots 0\\
    0 & 0 \cdots & 0 & 1& 1 \cdots &1 & & 0 & 0 & \cdots 0\\
    \vdots & \vdots & \vdots & \vdots& \vdots& \vdots & \cdots & \vdots & \vdots& \vdots \\
    0 & 0 \cdots & 0 & 0& 0 \cdots &0 & & 1 & 1 & \cdots 1\\
\end{array}
\right)_ {N \times N^2}
$.\\

The optimization problem in the vectorized form in Eq. (\ref{eq: y}), which includes four unequal constraints and a one-dimensional decision variable (tensor), is equivalent to Eq. (\ref{eq: original qp}). The three key properties of Eq. (\ref{eq: y}) are as follows:

\begin{itemize}
    \item The parameter from the prediction layer $ \hat{\mathbf{D}_f}^{\tau+1}$, is embedded exclusively in the objective function's term $q$ and does not appear in any of the constraint terms.
    \item The vectorized form simplifies the practical implementation of embedding the optimization layer within the neural network, allowing for a seamless connection with the prediction layer and the construction of the entire framework. 
    \item By proposing a generalized form with parameters represented as $G_n$ and $h_n$, we create a more straightforward representation of the optimization problem, facilitating the derivation of the solution algorithm later on. 
\end{itemize} 

\subsection{Aggregation}

This section integrates the actual distribution of DVs $\mathbf{D}_c^{\tau+1}$ and NDVs $\mathbf{D}_f^{\tau+1}$ obtained from optimization module to derive the holistic matching distribution of all vehicles $\mathbf{D}_a^{\tau+1}$ in the subsequent time interval $\tau + 1$. 
For DVs, by solving the optimization problem in Eq. (\ref{eq: y}), we can generate the optimal strategy for DVs. In particular, the actual allocated distribution of DVs $\mathbf{D}_c^{\tau+1}$  is then determined by aggregating the number of DVs that arrive in grids at time $\tau+1$:

\begin{equation}
\label{eq: D = sum y}
    \mathbf{D}_c^{\tau+1} = \mathbf{Ay}.
\end{equation}

For NDVs, these vehicles will reach their destinations by time $\tau + 1$, enabling the direct determination of their actual distribution, denoted as $\mathbf{D}_f^{\tau+1}$. Note that within the optimization module, we employ a surrogate distribution—specifically, the predicted distribution of NDVs $\hat{\mathbf{D}}_f^{\tau+1}$ at time $\tau+1$ to approximate the actual distribution $\mathbf{D}_f^{\tau+1}$. This predicted distribution serves as a proxy for $\mathbf{D}_f^{\tau+1}$ and plays a critical role in formulating the relocation plan.

Consequently, the actual holistic matching distribution for all vehicles can be straightforwardly represented by Eq. (\ref{eq: matching}):

\begin{equation}
\label{eq: matching}
    \mathbf{D}_c^{\tau+1} + \mathbf{D}_f^{\tau+1} = \mathbf{D}_a^{\tau+1}.
\end{equation}

\subsection{The integrated SPO model}

In the end-to-end SPO framework for the vehicle relocation problem in VCS, the primary objective is to enhance the sensing accuracy of the sensors. Since the sensors are installed on vehicles, the sensing distribution is effectively represented by the matching distribution of the vehicles. Consequently, the ultimate goal is to minimize the matching divergence between the matching distribution $\mathbf{D}_a^{\tau+1}$ and the target distribution $\mathbf{T}_a^{\tau+1}$ across all vehicles. The matching loss is formally defined as follows:

\begin{equation}
\label{eq: matching loss}
    \mathscr{L}_2 = \lVert {{\mathbf{T}_{a}^{\tau+1}} - \mathbf{D}_{a}^{\tau+1}} \rVert _2 ^2.
\end{equation}

Compared to $\mathscr{L}_1$ in Eq. (\ref{eq: prediction loss}), which is a mid-term prediction loss of NDVs, $\mathscr{L}_2$ in Eq. (\ref{eq: matching loss}) is the matching loss of all vehicles including DVs and NDVs.

From Eq. (\ref{eq: matching}), it is evident that both the prediction error and relocation error contribute to the overall matching divergence. Therefore, the loss function for the SPO framework is formulated by combining the prediction loss in Eq. (\ref{eq: prediction loss}) and the matching loss in Eq. (\ref{eq: matching loss}):

\begin{equation}
\label{eq: SPO loss}
    \mathscr{L}_{SPO} = w_1 \mathscr{L}_1 + w_2 \mathscr{L}_2,
\end{equation}
where $w_1$ is the weight for prediction loss for the NDVs, and $w_2$ is the weight for the matching loss for all vehicles.

As illustrated in Figure \ref{fig: end to end}, the vehicle relocation problem defined in Eq. (\ref{eq: y}) is embedded in the optimization module and connected with the preceding prediction module in Eq. (\ref{eq: tgcn model}) via gradients. This integration allows the prediction module to inform the optimization process through the backpropagation of gradients, facilitating end-to-end training. Consequently, the SPO framework can be formally presented as follows:

\begin{mini}{}
    {Eq.\ (\ref{eq: SPO loss})} 
    {\label{eq: integrated model}}{}
    \addConstraint{Eq.\ (\ref{eq: tgcn model})}{ -\text{Prediction }}
    \addConstraint{Eq.\ (\ref{eq: y})}{ -\text{Vehicle relocation problem }}
    \addConstraint{Eq.\ (\ref{eq: D = sum y}, \ref{eq: matching})}{-\text{Aggregation.}}
\end{mini}

Eq. (\ref{eq: integrated model}) has two crucial properties as summarized below:

\begin{itemize} 
    \item This is a sequential decision-making problem where the solution process progresses from addressing the lower-level constraints to solving the upper-level minimization function. It begins with prediction in Eq. (\ref{eq: tgcn model}), followed by optimization in Eq. (\ref{eq: y}), and concludes with the calculation of divergence in Eq. (\ref{eq: SPO loss}).
    
    \item The overall nested optimization problem contains a sub-constrained optimization problem (\ref{eq: y}) within the constraint section, making it challenging to solve directly and efficiently.
\end{itemize}

Considering these properties, it is essential to develop novel solution methods to effectively address such nested-constrained optimization problems within the deep learning architecture.

\subsection{Solution algorithm}

In this section, we will first introduce the computational graph in the SPO framework, then propose the explicit unrolling approach of the SPO framework, including the forward pass by the ADMM, and the backward alternating differentiation method.

\subsubsection{The computational graph}

\begin{figure}[h]
    \centering
    \includegraphics[width=0.9\textwidth]{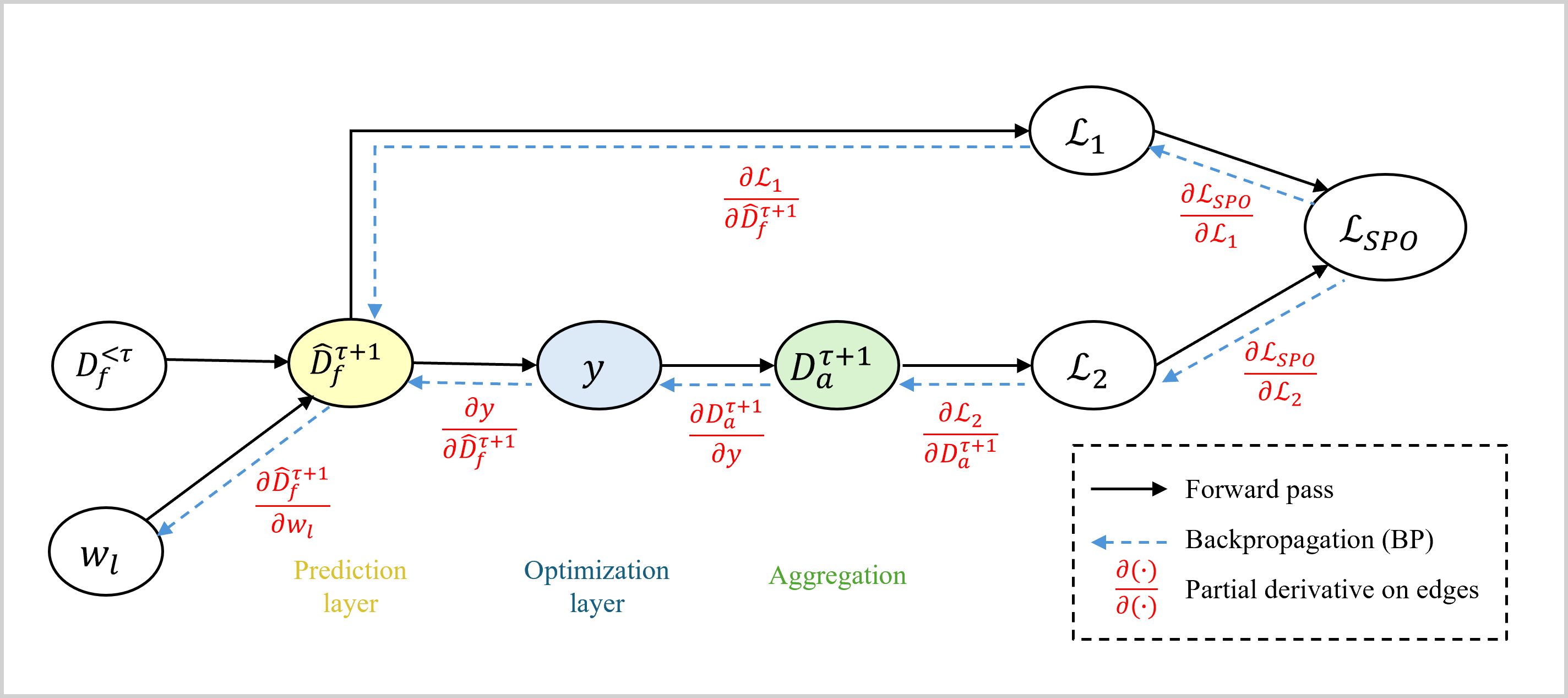}
    \caption{Illustration of the computational graph.}
    \label{fig: computational graph}
\end{figure}

The computational graph illustrates how the gradients link and propagate in the neural network in the forward and backward pass. Therefore, we first propose the computational graph in the SPO framework, as is shown in Figure \ref{fig: computational graph}.
The computation of gradients is essential to enable backpropagation (BP) within the neural network. illustrates the forward and backward pass process in the SPO framework.

To solve the SPO framework as formulated in Eq. (\ref{eq: integrated model}), we derive the gradients of the loss function $\mathscr{L}_{SPO}$ w.r.t. the weight $\mathbf{w}_p$ in the prediction module, as is shown in Eq. (\ref{LSPO derivative}):

\begin{align}
\label{LSPO derivative}
    \frac{\partial \mathscr{L}_{SPO}}{\partial \mathbf{w}_p}
    &=\frac{\partial \mathscr{L}_{SPO}}{\partial\mathscr{L}_{1}} 
      \frac{\partial \mathscr{L}_{1}}{\partial\hat{\mathbf{D}}_f^{\tau+1}}
      \frac{\partial \hat{\mathbf{D}}_f^{\tau+1}}{\partial \mathbf{w}_p} 
      + 
      \frac{\partial \mathscr{L}_{SPO}}{\partial\mathscr{L}_{2}} 
      \frac{\partial \mathscr{L}_{2}}{\partial \mathbf{D}_a^{\tau+1}}
      \frac{\partial \mathbf{D}_a^{\tau+1}}{\partial \mathbf{y}}
      \frac{\partial \mathbf{y}}{\partial \mathbf{\hat{D}}_f^{\tau+1}}
      \frac{\partial \hat{\mathbf{D}}_f^{\tau+1}}{\partial \mathbf{w}_p}. 
\end{align}

All the derivatives can be computed analytically and automatically by the Autograd package in PyTorch, except for the $\frac{\partial \mathbf{y}}{\partial \hat{\mathbf{D}_f}^{\tau+1}}$ in the optimization layer. Hence, the primary challenge is to compute the gradient of the optimization layer to enable forward and backward propagation. To this end, the gradient of the optimization layers necessitates solving the internal optimization problem first and subsequently differentiating the solution with respect to the parameter provided from the preceding layer. 

\subsubsection{The unrolling K-layer neural network}

\begin{figure}[h]
    \centering
    \includegraphics[width=0.95\textwidth]{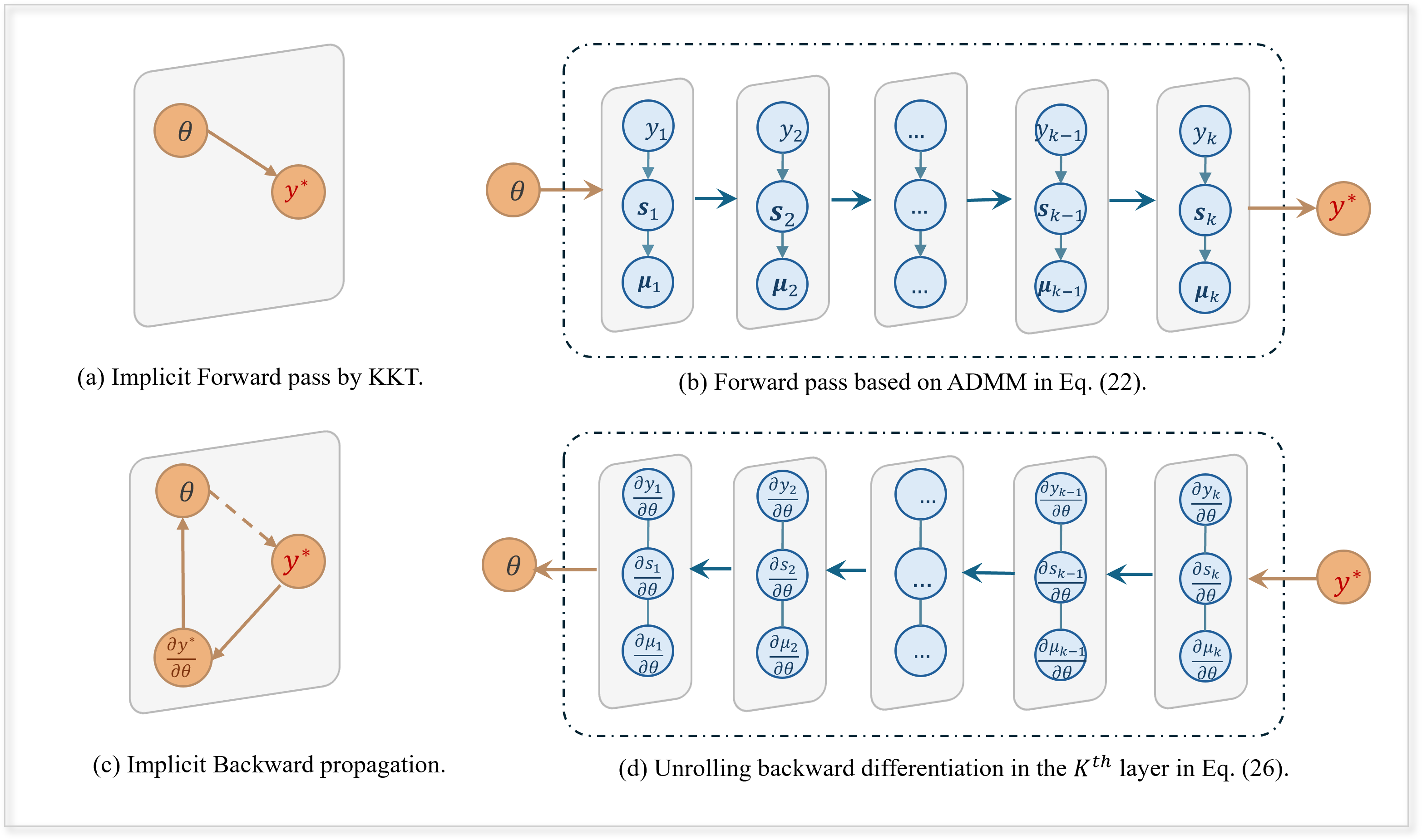}
    \caption{A comparison of the explicit and implicit differentiation method. (a) and (c) show the forward and backward pass through the implicit differentiation method by KKT, (b) and (d) depict the forward and backward pass through the alternating differentiation method. }
    \label{fig: unrolling}
\end{figure}

Then we proceed to introduce how to derive the gradient of the optimal solution with respect to the predefined parameter from the prediction layer, $\frac{\partial \mathbf{y}}{\partial \hat{\mathbf{D}_f}^{\tau+1}}$, using an unrolling approach through the alternating differentiation method. 
Figure \ref{fig: unrolling} illustrates the unrolling alternating differentiation method and compares it with the implicit differentiation method. 

In the implicit differentiation method by KKT condition in Figure \ref{fig: unrolling} (a) and (c), the solution for forward is obtained directly, and the corresponding gradient for the backward pass is computed in a single layer. However, obtaining these gradients directly becomes challenging in large-scale networks due to the high computational cost associated with calculating the Jacobian matrix in the KKT condition.

To address this, we propose an alternating differentiation approach, as shown in Figure \ref{fig: unrolling} (b) and (d). This approach unfolds the one-layer network into $K$ iterations through an alternating updating procedure. In the forward pass, the problem is solved using the ADMM, where the primal, slack, and dual variables are computed iteratively over $K$ iterations until convergence (or when the optimal solution is found). Concurrently, the gradients for these variables are computed simultaneously through analytical functions at each step. Each iteration in the forward pass is mapped to a single layer. Once convergence is achieved, these $K$ layers are stacked together, forming a $K$-layer neural network. The final gradient in the optimizer for the alternating differentiation method is obtained by intrinsically combining the gradients from all $K$ layers. Then, in the backward pass, the final gradient $\frac{\partial \mathbf{y}}{\partial \hat{\mathbf{D}_f}^{\tau+1}}$ directly propagates at $y^*$ in the final $K^{th}$ layer, which inherently corresponds to passing through all iterating gradients. Essentially, the unrolled $K$-layer neural network is equivalent to a one-layer neural network used in the implicit differentiation method.

Then we will introduce the solution algorithm for the forward pass and backward pass of the unrolled alternating differentiation method, respectively.

{\bf Forward pass based on the ADMM.}
The optimization problem in Eq. (\ref{eq: y}) is a constrained quadratic problem with four unequal constraints. It can be solved using the ADMM by transforming the original constrained problem into an unconstrained optimization problem. This section derives the forward pass and the optimal solution via the ADMM.  

Note that the parameter from the prediction layer $\hat{\mathbf{D}_f}^{\tau+1}$ is embedded in the objective function's term $q(\hat{\mathbf{D}_f}^{\tau+1})$. Here we replace $\hat{\mathbf{D}_f}^{\tau+1}$ as $\theta$ as simplicity, and $q(\hat{\mathbf{D}_f}^{\tau+1})$ is replaced by $q(\theta)$. 

Following the procedures of the ADMM, we first form the augmented Lagrange function of the problem in Eq. (\ref{eq: y}) with a quadratic penalty term as follows:

\begin{equation}
    \begin{aligned}{\label{eq: aug_lag}}
        \max_\mathbf{\mu} \min_{\mathbf{y}\geq 0, \mathbf{s} \ge 0} \mathcal{L}(\mathbf{y}, \mathbf{s}, \mathbf{\mu}; \theta) 
        &= \frac{1}{2}\mathbf{y}^{\prime}\mathbf{P}\mathbf{y} + \mathbf{q(\theta)}^{\prime}\mathbf{y} 
        + \sum_{n=1}^4 \langle \mathbf{\mu}_n, \; \mathbf{G}_n\mathbf{y} + \mathbf{s}_n - \mathbf{h}_n \rangle \\
        &+ \frac{\rho}{2} (\sum_{n=1}^4 {\lVert \mathbf{G}_n\mathbf{y} + \mathbf{s}_n - \mathbf{h}_n\rVert}^2),
    \end{aligned}
\end{equation}
where $ \mathbf{s} = \left \{ \mathbf {s}_n|n = 1,2,3,4 \right\} \ge 0$ is the set of non-negative slack variables corresponding to the four constraints in Eq. (\ref{eq: y}), $ \mathbf{\mu} = \left\{\mathbf{\mu}_n| n = 1,2,3,4 \right\}$ is the set of dual variables of the four constraints, $\rho>0$ is the penalty term, $\theta$ is the parameter from the prediction layer and $\theta =  \hat{\mathbf{D}_f}^{\tau+1}$. 

Then the primary, slack, and dual variables are updated alternatively based on the ADMM.

\begin{subnumcases} {\label{eq: update y s mu}}
    \mathbf{y}_{k+1} = \argmin_{\mathbf{y}} \mathcal{L} (\mathbf{y}, \mathbf{s}_{n,k}, \mathbf{\mu}_{n,k}; \theta)\quad \forall n \in \left\{1, 2, 3, 4 \right\}, \label{eq: update y_k} \\
    \mathbf{s}_{n, k+1} = \argmin_{\mathbf{s}_n \geq 0} \mathcal{L}(\mathbf{y}_{k+1}, \mathbf{s}_{n,k}, \mathbf{\mu}_{n,k}; \theta)  \quad \forall n \in \left\{1, 2, 3, 4 \right\}, \label{eq: update s_k}\\
    \mathbf{\mu}_{n, k+1} = \mathbf{\mu}_{n, k} + \rho (\mathbf{G}_n \mathbf{y}_{k+1} + \mathbf{s}_{n, k+1} - \mathbf{h}_n) \quad \forall n \in \left\{1, 2, 3, 4 \right\}. \label{eq: update mu_k}
\end{subnumcases}

Then we derive the analytical updating solutions for the primal $\mathbf{y}_{k+1}$, slack $\mathbf{s}_{n, k+1}$, and dual variables $\mathbf{\mu}_{n, k+1}$, respectively.

Taking the first and second-order derivative of Eq. (\ref{eq: aug_lag}) w.r.t. $\mathbf{y}$ and we have $\nabla_\mathbf{y} \mathcal{L}$ and $\nabla^2_\mathbf{y} \mathcal{L}$ as follows: 

\begin{align}
    \nabla_\mathbf{y} \mathcal{L} &= 
    \mathbf{P}^{\prime}\mathbf{y} + \mathbf{q(\theta)} + \mathbf{G}_n^{\prime} \mathbf{\mu}_n+ \sum_{n=1}^4 \rho (\mathbf{G}_n^{\prime} (\mathbf{G}_n\mathbf{y} + \mathbf{s}_n - \mathbf{h}_n)) {\label{eq: first_order_of_L_to_y}}, \\
    \nabla^2_\mathbf{y} \mathcal{L}(\mathbf{y}_{k+1}) &= {\mathbf{P} + \sum_{n=1}^4\rho \mathbf{G}_n^{\prime} \mathbf{G}_n} {\label{eq: second order of L to y}}.
\end{align}

Let Eq. (\ref{eq: first_order_of_L_to_y}) equals $0$ and $\mathbf{y}_{k+1}$ in Eq. (\ref{eq: update y_k}) is therefore obtained as follows:

\begin{equation}{\label{eq: y by first order}}
    \begin{aligned}
{-(\mathbf{P} + \sum_{n=1}^4 \rho \mathbf{G}_n^{\prime} \mathbf{G}_n)} \mathbf{y}_{k+1}  = (\mathbf{q(\theta)} + \sum_{n=1}^4 \rho \mathbf{G}_n^{\prime} (\mathbf{s}_{n,k} - \mathbf{h}_n) + \sum_{n=1}^4 \mathbf{G}_n^{\prime} \mathbf{\mu}_{n,k}). 
    \end{aligned}   
\end{equation}

Let $U = - {(\mathbf{P} + \sum_{n=1}^4 \rho \mathbf{G}_n^{\prime} \mathbf{G}_n)}^{\prime}$ and $U$ is constant. 
Then $\mathbf{y}_{k+1}$ can be updated by Eq. (\ref{eq: y with U}):

\begin{equation}{\label{eq: y with U}}
    \begin{aligned}
     \mathbf{y}_{k+1}  & = U 
     (\mathbf{q}(\theta) + \sum_{n=1}^4 \rho (\mathbf{G}_n^{\prime} (\mathbf{s}_{n,k} - \mathbf{h}_n) + \mathbf{G}_n^{\prime} \mathbf{\mu}_{n,k})). 
    \end{aligned}   
\end{equation}

Taking the first-order derivative of Eq. (\ref{eq: aug_lag}) w.r.t. $\mathbf{s}_{n,k}, \forall n \in \left\{ 1,2,3,4\right\}$ , the slack variables $\mathbf{s}_{n, k+1}$ in Eq. (\ref{eq: update s_k}) are substituted by Eq. (\ref{eq: skn_by_relu}) with a ReLU function:

\begin{equation}{\label{eq: skn_by_relu}}
    \begin{aligned}
        \mathbf{s}_{n,k+1} = \mathbf{ReLU}( - \frac{1}{\rho} \mathbf{\mu}_{n,k} - (\mathbf{G}_n\mathbf{y}_{k+1} - \mathbf{h}_n)).
    \end{aligned}
\end{equation}

We finally obtain the forward solution of primal, slack, and dual variables in Eq. (\ref{explicit ysmu}):

\begin{subnumcases}{\label{explicit ysmu}}
    \mathbf{y}_{k+1} = U(\mathbf{q}(\theta) + \sum_{n=1}^4 \rho \mathbf{G}_n^{\prime} (\mathbf{s}_{n,k} - \mathbf{h}_n) + \sum_{n=1}^4 \mathbf{G}_n^{\prime} \mathbf{\mu}_{n,k} \label{eq: yk}),\\
    \mathbf{s}_{n,k+1} = \mathbf{ReLU}( - \frac{1}{\rho} \mathbf{\mu}_{n,k} - (\mathbf{G}_n\mathbf{y}_{k+1} - \mathbf{h}_n)) \quad \forall n \in \left\{1, 2, 3, 4 \right\},\label{eq: sk}\\
    \mathbf{\mu}_{n, k+1} = \mathbf{\mu}_{n, k} + \rho (\mathbf{G}_n \mathbf{y}_{k+1} + \mathbf{s}_{n, k+1} - \mathbf{h}_n) \quad \forall n \in \left\{1, 2, 3, 4 \right\}. \label{eq: muk}
\end{subnumcases}

{\bf Backward pass by the alternating differentiation method.}
\label{sec: differentiation algorithm}
Based on the forward solving procedures in Eq. (\ref{explicit ysmu}), we can then present the backward differentiation algorithm, which is to compute the derivatives of the primal, slack, and dual variables with respect to predefined parameters alternatively to enable backpropagation.

In the end-to-end SPO framework, we address a real-time vehicle relocation problem in Eq. (\ref{eq: y}), where the pre-defined parameter (output) from the prediction layer is denoted as $\hat{\mathbf{D}}_f^{\tau+1}$. In this section, we regard $\hat{\mathbf{D}}_f^{\tau+1}$ as $\theta$ for simplicity, and $\theta$ is in $\mathbf{q(\theta)}$.

Applying the Implicit Function Theorem \citep{krantz2002implicit}  to Eq. (\ref{eq: first_order_of_L_to_y}), then the derivative of the solution $\mathbf{y}^{k+1}$ w.r.t. the parameter $\theta$ can be formulated as:
\begin{align}
    \frac{\partial{\mathbf{y}^{k+1}}}{\partial{\theta}} & = - \nabla^2_\mathbf{y} \mathcal{L}(\mathbf{y}_{k+1})^{\prime} {\nabla_{\mathbf{y}, \theta} \mathcal{L}(\mathbf{y}_{k+1})} = U  (\frac{\partial{\mathbf{q}(\theta)}}{\partial \theta} + \sum_{n=1}^4 \rho \mathbf{G}_n^{\prime} \frac{\partial \mathbf{s}_{n,k}}{\partial \theta} + \sum_{n=1}^4 \mathbf{G}_n^{\prime} \frac{\partial \mathbf{\mu}_{n,k}}{\partial \theta}),
\end{align}
where the second-order derivative $\nabla^2_\mathbf{y} \mathcal{L}$ is obtained in Eq. (\ref{eq: second order of L to y}).

The derivative of $\mathbf{s}_{n, k+1}$ w.r.t. $\theta$ can be obtained as follows:

\begin{equation}
\label{eq: s to theta}
    \frac{\partial{\mathbf{s}_{n, k+1}}}{\partial \theta}  = - \frac{1}{\rho} \mathbf{sgn}(\mathbf{s}_{n, k+1}) \cdot \mathbf{1}^{\prime} \odot (\frac{\partial \mathbf{\mu}_{n,k}}{\partial \theta} + \rho \frac{\partial(G\mathbf{y}_{n, k+1} - \mathbf{h}_n)}{\partial \theta}).
\end{equation}

The derivative of $\mathbf{\mu}_{n, k+1}$ w.r.t. $\theta$ can be easily obtained as follows:

\begin{equation}
    \frac{\partial{\mathbf{\mu}_{n, k+1}}}{\partial \theta} = \frac{\partial \mathbf{\mu}_{n,k}}{\partial \theta} + \rho \frac{\partial(G\mathbf{y}_{n, k+1} + \mathbf{s}_{n, k+1} - \mathbf{h}_n)}{\partial \theta}.  
\end{equation}

We finally obtain the explicit differentiation function of primal, slack and dual variables as summarized in Eq. (\ref{eq: explicit diff}):

\begin{subnumcases}{\label{eq: explicit diff}}
    \frac{\partial{\mathbf{y}_{k+1}}}{\partial \theta}  = U  (\frac{\partial{\mathbf{q}(\theta)}}{\partial \theta} + \sum_{n=1}^4 \rho \mathbf{G}_n^{\prime} \frac{\partial \mathbf{s}_{n,k}}{\partial \theta} + \sum_{n=1}^4 \mathbf{G}_n^{\prime} \frac{\partial \mathbf{\mu}_{n,k}}{\partial \theta}),  \label{eq: y_k to theta} \\
    \frac{\partial{\mathbf{s}_{n, k+1}}}{\partial \theta}  = - \frac{1}{\rho} \mathbf{sgn}(\mathbf{s}_{n, k+1}) \cdot \mathbf{1}^{\prime} \odot (\frac{\partial \mathbf{\mu}_{n,k}}{\partial \theta} + \rho \frac{\partial(\mathbf{G}_n\mathbf{y}_{ k+1} - \mathbf{h}_n)}{\partial \theta}), \label{eq: s_k to theta} \\
    \frac{\partial{\mathbf{\mu}_{n, k+1}}}{\partial \theta} = \frac{\partial \mathbf{\mu}_{n,k}}{\partial \theta} + \rho \frac{\partial(\mathbf{G}_n\mathbf{y}_{ k+1} + \mathbf{s}_{n, k+1} - \mathbf{h}_n)}{\partial \theta}, \label{eq: mu_k to theta}
\end{subnumcases}
where 
$ \frac{\partial q}{\partial \theta} = \mathbf{A}^{\prime}$ and $\odot$ is the Hadamard product. In particular, Eq. (\ref{eq: explicit diff}) is utilized to perform the backward propagation on the computational graph. 
Note the above forward pass by the ADMM in Eq. (\ref{explicit ysmu}) and the backward differentiation in Eq. (\ref{eq: explicit diff}) is suitable for any quadratic problems with unequal constraints in the form in Eq. (\ref{eq: y}).

\section{Solution algorithms}
\label{sec: pseudo code}

The overall solution algorithm of the proposed SPO framework is presented in Algorithm \ref{alg: whole procedure}, which can be performed in batch form.

\begin{algorithm}[h]
\caption{Solving procedure for the end-to-end SPO framework.}
\label{alg: whole procedure}
    \SetKwProg{Fn}{Function}{:}{}
    \SetKwFunction{Optimization}{Optimization}
    \KwData{Input the historical demand $\mathbf{D}_f^{\leq\tau}$, the target distribution $\mathbf{T}_a^{\tau+1}$, the control ratio $\gamma$, the total incentive cost  budget $R$, the actual travel time matrix $C$, the penalty term $\rho$, the convergence threshold $\xi$.}
    \KwResult{Output the relocation strategy $\mathbf{y}^*$, the matching distribution $\mathbf{D}_a$.}
    Initialize Max epoch $N_e$, prediction weight $\mathbf{w_1}$, matching weight $\mathbf{w_2}$ \;
    \For {\textnormal{Epoch} $\ e = 1,..., N_e$}{
        Input the $\mathbf{D}_f^{\leq \tau}$ into prediction module \; 
        Predict the distribution of NDVs $\hat{\mathbf{D}}_f^{ \tau+1}$ in Eq. (\ref{eq: tgcn model}) \;
        Compute prediction loss $\mathscr L_1$ in Eq. (\ref{eq: prediction loss}) 
        Input $\mathbf{\hat{D}}_f^{ \tau+1}$ and $\mathbf{D}_c^{\tau}$ into optimization module in Eq. (\ref{eq: y}) \;
        Solve \Optimization{} \;
        \Fn{\Optimization{}}{
            Input the parameter $\theta$ \;
            Initialize $k \gets 0$ \;
            Initialize variables and derivatives $\mathbf{y}_k, s_{i,k}, \mathbf{\mu}_{i,k},
                        \frac{\partial{\mathbf{y}_{k}}}{\partial \theta}, 
                        \frac{\partial{\mathbf{s}_{i,k}}}{\partial \theta}, 
                        \frac{\partial{\mathbf{\mu_i}_{k}}}{\partial \theta}$ \;
            Initialize parameters $\mathbf{G_1, G_2, G_3, G_4}, \mathbf{h_1, h_2, h_3, h_4}, \mathbf{A}, \mathbf{B}$, convergence threshold $\xi$\ \;
            \While{$ |\hat{Z}_{k+1} - Z_k | \geq \xi$ }{ 
                Primal update $\mathbf{y}_k$ by Eq. (\ref{eq: yk}), $\frac{\partial{\mathbf{y}_{k}}}{\partial \theta}$ by Eq. (\ref{eq: y_k to theta}) \;
                Slack update $s_{n,k}$ by Eq. (\ref{eq: sk}), $\frac{\partial{s_{n,k}}}{\partial \theta}$ by Eq. (\ref{eq: s_k to theta}) \;
                Dual update $\mathbf{\mu}_{n,k}$ by Eq. (\ref{eq: muk}), $\frac{\partial{\mathbf{\mu}_{n,k}}}{\partial \theta}$ by Eq. (\ref{eq: mu_k to theta}) \;
                Compute $\hat{Z}_k$ \;
                $k \gets k+1 $ \;
            }
            \Return the relocation strategy $\mathbf{y}^\ast$ \;
        }
        Aggregate $\mathbf{y}^\ast$ and obtain distribution for DVs $\hat{D}_c^{\tau+1}$ in Eq. (\ref{eq: D = sum y}) \;
        Obtain matching distribution $\hat{\mathbf{D}}_a^{\tau+1}$ in Eq. (\ref{eq: matching}) \;
        Compute the SPO loss $\mathscr L_2$ in Eq. (\ref{eq: SPO loss})\;
        Back propagate and update weights $w_p$ in prediction model\;
    }
\end{algorithm}

\section{Numerical experiments}
\label{sec: experiment}

In this section, we evaluate the proposed end-to-end SPO framework embedded with an unrolling approach using real-world taxi data from two distinct locations: the Kowloon district in Hong Kong SAR and Chengdu City, China. First, we outline the experiment settings of the two cases and then present the experimental results to verify the effectiveness and efficiency of the proposed framework in each case. For both cases, we compare the proposed SPO-A across mid-size and large-scale datasets to assess scalability against various baseline models, including the SPO-C framework and the conventional two-stage PTO framework. In Case A, we perform ablation studies and sensitivity analyses to examine the influence of key parameters on the proposed SPO-A framework. In Case B, we extend our evaluation to larger-scale parallel experiments, further validating the robustness and applicability of the proposed framework in a larger-scale urban setting. All the experiments are conducted on a desktop with Intel Core i7-13700K CPU 3.40 GHz $\times$ 32G RAM, 500 GB SSD, GeForce RTX 3090 Ti GPU.

\subsection{General settings}

In this section, we introduce the characteristics of the research areas, hyper-parameter configuration, baseline models, target sensing distributions, and evaluation metrics for the two cases.

\subsubsection{Experimental setup}
Table \ref{tab: experiment} summarizes the detailed experimental setups for the two cases. Both experiments are conducted using real-world road networks and taxi datasets. Compared to Case A, Case B encompasses a broader geographical region and a larger network scale. Additionally, the average taxi demand per grid per time interval in Case B is roughly three times higher than in Case A. In both cases, the ratio of DVs to NDVs varies from 20\% to 80\% with an increment of 10\%. While the prediction module in Case A only incorporates historical taxi demand as input features, Case B employs a more comprehensive feature set, including historical demand, weather conditions, and the number of Points of Interest (POIs).

\begin{table}[h]
    \caption{Experimental setup in Case A and Case B.}
    \label{tab: experiment}
    \centering
    \addtolength{\tabcolsep}{5pt}
    \begin{tabular}{ccc}
        \toprule
        & Case A & Case B \\
        \midrule
        Area size & 50.11 $km^2$ & 73.73 $km^2$ \\
        Total data sample & 2,075,864 & 9,985,238 \\
        Average demand in each grid & 35 & 126\\      
        Network scale & $45 \times 45, 68\times 68$ & $60 \times 60, 80\times 80, 100\times 100$\\
        Prediction features & Historical demand & Historical demand, weather, POI \\
        \bottomrule
    \end{tabular}
\end{table}

\subsubsection{Hyper-parameter setting}

The hyper-parameters in the SPO framework of Case A and Case B are summarized in Table \ref{tab: hyperparameter}. We utilize the same Adagrad optimizer, weight decay, look-back window, and time interval for both cases. However, since Case B involves a larger-scale dataset than Case A  and requires significantly longer training times, we adopt a higher learning rate, a more relaxed convergence threshold, and a stronger penalty term to accelerate convergence while balancing accuracy and computational efficiency. Additionally, we use a small batch size of 16 for Case B to avoid memory overflow during the computation of the large Jacobian. All hyper-parameters in both the prediction and optimization modules of the two cases are carefully fine-tuned and selected through cross-validation to ensure optimal performance. The setting of prediction modules is provided in \ref{sec: appendix B} (See Table \ref{apd: hyperparameter}). The cross-validation procedure of the penalty term is supplemented in Section \ref{sec: penalty term}.

\redrevise{
\begin{table}[h]
    \centering
    \caption{Configuration of the SPO framework in Case A and Case B.}
    \addtolength{\tabcolsep}{10pt}
    \begin{threeparttable}
    \begin{tabular}{ccc}
        \toprule
        Hyper-parameters  &  Case A & Case B\\
        \midrule
        Optimizer & Adagrad & Adagrad \\
        Batch size & 64 & 16\\
        Learning rate & $\text{5e-3}$ & $\text{1e-2}$\\
        Weight decay & $\text{1e-4}$ & $\text{1e-4}$\\
        Look-back window & 12 & 12\\
        Time interval & 15 min & 15 min \\
        Convergence threshold & $\text{5e-2}$ & $1$ \\
        Penalty term in optimization & 2.0 & 5.0 \\
        Total travel budget for optimization & $8,000/10,000/12,000/15,000$ & $30,000$\\
        Target distributions & U,\ G,\ GM \tnote{a} & U, G, GM \tnote{b}\\
        \bottomrule
    \end{tabular}
    \begin{tablenotes}
    \footnotesize
    \item[a,b] The parameters in U, G, and GM for Case A and Case B are different.
    \end{tablenotes}   
    \end{threeparttable}
    \label{tab: hyperparameter}
\end{table}
}

\subsubsection{Dynamic weighting strategies}

This section summarizes the tuning techniques applied in the proposed SPO framework to prevent overfitting. While standard methods such as dropout and learning rate scheduling are employed, we additionally adopt two innovative techniques specific to the SPO framework: the dynamic weighting mechanism applicable to both SPO-A and SPO-C frameworks, and optimization layer regularization via the penalty term to the SPO-A. 

\begin{itemize}
    \item \textit{Dynamic weight adjustment and warm initialization.} The final SPO loss function provided in Equation \ref{eq: SPO loss} combines both the prediction MSE and optimization MSE. To harmonize their convergence, we implement a phased weighting strategy: 1) Initial Phase (Epochs 1–100): Prediction loss weight is amplified 20–50 times over optimization loss to stabilize early training via warm initialization. 2) Transition Phase (Next 100 Epochs): Prediction weight is linearly reduced to achieve a 1:1 balance, ensuring equitable contributions from both prediction and optimization. 3) Final Phase (Last few epochs): The ratio is either maintained at 1:1 or slightly tilted towards the optimization loss to refine decision quality. The weight setting procedure is provided in Table \ref{tab: weight setting}.
    \item \textit{Optimization layer regularization via ADMM penalty tuning.} In the SPO-A framework, the penalty term $\rho$ serves a dual role: an adaptive learning rate for dual variable updates while simultaneously transforming the original hard constraints into a more flexible soft solution space. By carefully tuning $\rho$ through cross-validation, we balance the trade-off between constraint satisfaction and model flexibility, enhancing generalization without compromising feasibility. The value of the penalty term is selected through cross-validation, which is provided in Section \ref{sec: penalty term}.
\end{itemize} 

\begin{table}[h]
    \centering
    \caption{Dynamic weight setting of prediction to matching MSE in Case A and Case B. }
    \label{tab: weight setting}
    \begin{threeparttable}
    \addtolength{\tabcolsep}{10pt}
    \begin{tabular}{ccc}
    \toprule
       Epoch  & Case A & Case B \\
    \midrule
        $0 \thicksim 100 $& $50:1$ & $100:1$\\
        $100\thicksim200$ & $50:1 \to 1:1$\tnote{a} & $100:1 \to 1:1$\tnote{b} \\
        $> 200$ & $1:1$ & $1:1.05$ \\
    \bottomrule
    \end{tabular}
    \begin{tablenotes}
    \footnotesize
    \item[a,b] The ratio is scheduled to decrease linearly over each epoch.
    \end{tablenotes}       
    \end{threeparttable}
\end{table}

These two techniques offer complementary advantages for managing the prediction-optimization balance and ensuring solution feasibility for the SPO framework. When integrated with conventional regularization methods, they collectively form a comprehensive approach that simultaneously prevents overfitting while preserving high-quality decision quality across diverse operational scenarios.

\subsubsection{Baselines}

To compare the importance of both the SPO framework and the embedded alternating differentiation method (SPO-A), we compare the proposed model with three baseline methods: SPO with CVXPY (SPO-C), 2-stage predict-then-Optimize (PTO), and do-nothing method (DON). 

\begin{itemize}
    \item SPO-C: The SPO-C is an end-to-end SPO framework embedded with the implicit differentiation algorithm CVXPY by differentiating the KKT condition.
    \item PTO: The PTO method is the two-stage predict-then-optimize framework, in which the matching strategy is generated completely based on the prediction result.
    \item DON: The DON method does not change the original routes of the DVs and lets the drivers drive by themselves.
\end{itemize}

\subsubsection{Target sensing distributions}

The target sensing distribution can vary significantly depending on specific sensing tasks in VCS, necessitating that the proposed SPO-A framework exhibit robustness across a wide range of target sensing distributions. For instance, air quality monitoring typically requires uniformly distributed data collection across an entire urban area \citep{bales2012citisense, chen2018pga}. In contrast, specialized monitoring tasks, such as detecting factory pollution or forest fires, often demand more frequent and granular data collection in densely populated regions or specific locations during particular dates or seasons \citep{khedo2010wireless, paulos2007sensing}.

Therefore, we create three different target distributions: Uniform distribution, Gaussian distribution, and Gaussian Mixture distribution \citep{xu2019ilocus} based on the one-week data from March 7 to March 13 to represent various potential requirements. Each distribution is generated using unique random seeds to ensure reproducibility while maintaining statistical independence. The specific generation procedures are as follows:

\begin{itemize}
    \item $\textbf{\textit{Uniform distribution}} \ (U)$. We first compute the mean distribution of the one-week data in all grids for each time interval. We then generate uniformly distributed samples within the range $[0.90, 1.10]$ and $[0.85, 1.15]$ of the mean value for each grid for Case A and Case B, respectively. Negative values are filtered.
    \item $\textbf{\textit{Gaussian distribution}} \ (G)$. Based on the mean distribution for each time interval, we generate samples from a Gaussian distribution across the spatial domain. The distribution is centered at the mean with a fixed variance (15 for Case A and 25 for Case B). Negative values are filtered.
    \item $\textbf{\textit{Gaussian Mixture distribution}} \ (GM)$. We sample a Gaussian Mixture distribution over the spatial domain based on the mean distribution with two components of two different fixed variances for each time interval (10 and 20 for Case A, 40 and 50 for Case B, respectively). Negative values are filtered.
\end{itemize}

\subsubsection{Evaluation metrics}

To evaluate the matching performance of the SPO framework, we adopt two metrics to compare the divergence between the matching distribution $\mathbf{D}_a^{\tau+1}$ and the target distribution $\mathbf{T}_a^{\tau+1}$. Since the sensors are installed on vehicles, enhanced matching performance reflects better sensing accuracy for a given sensing task.

\begin{itemize}
    \item \textbf{\textit{Root Mean Squared Error (RMSE)}}:\\
    $\text{RMSE} = \sqrt{\frac{1}{N}\sum_{i=1}^{N}(\mathbf{D}_{a,i}^{\tau+1} - \mathbf{T}_{a,i}^{\tau+1})^2}$
    \item \textbf{\textit{Symmetric Mean Squared Percentage Error (SMAPE)}}:\\
    $\text{SMAPE} = \frac{100\%}{N}
    \sum_{i=1}^{N}\frac{|\mathbf{D}_{a,i}^{\tau+1} - \mathbf{T}_{a,i}^{\tau+1}|}{(|\mathbf{D}_{a,i}^{\tau+1} + |\mathbf{T}_{a,i}^{\tau+1}|)/2}$ \\
\end{itemize}

\subsection{Case A: Mid-size hailing dataset}

\label{subsec: research area}
\begin{figure}[h]
    \centering
    \includegraphics[width=0.95\textwidth]{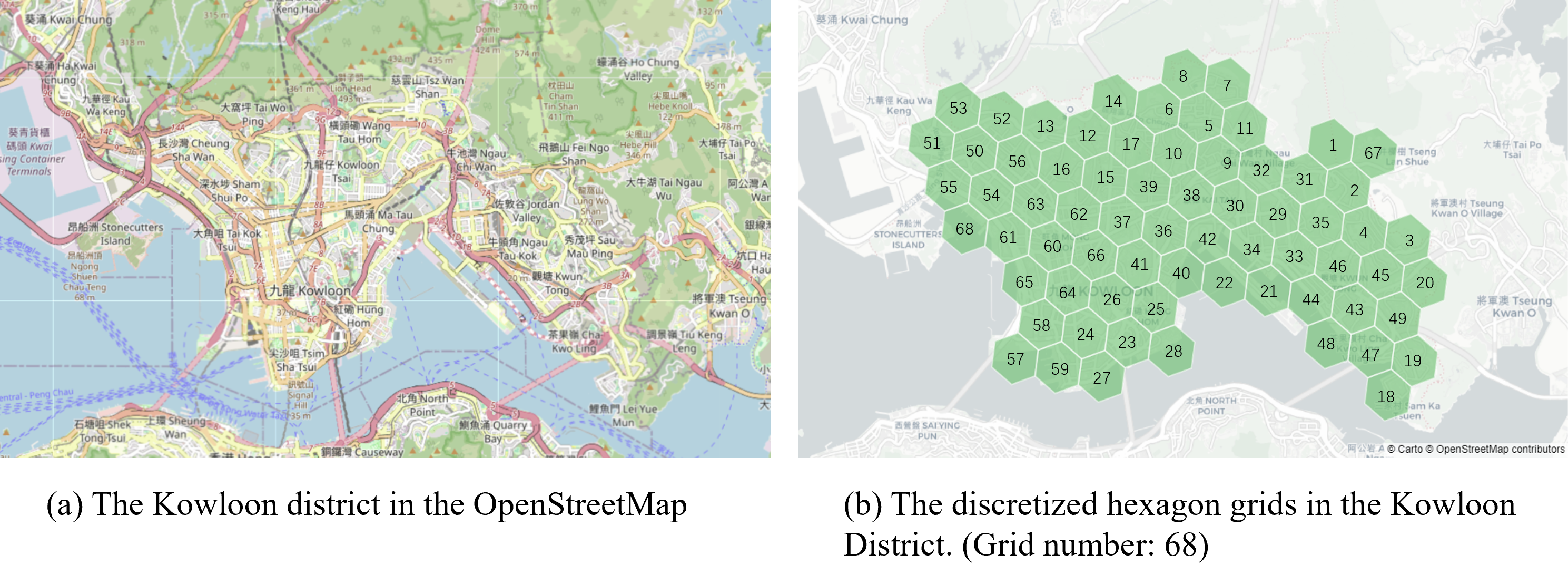}
    \caption{Overview of the research area in the Kowloon District, Hong Kong SAR.}
    \label{fig: research area}
\end{figure}

In this section, we conduct the first experiment in the real-world taxi dataset in Hong Kong. The study area in the Kowloon District, Hong Kong SAR, is first discretized into hexagon grids by Uber's Hexagonal Hierarchical Spatial Index (H3) \citep{h3} at H8 resolution. The average edge length of each hexagon is 531.41 $m$. For the mid-size network, we select 45 grids, and for the large-scale network, 68 grids are chosen, as illustrated in Figure \ref{fig: research area}. Real-time taxi information is collected from the HKTaxi application programming interface (API) from March 14 to March 24, 2023, and the taxi demand in each grid is aggregated every 15 minutes. Following preprocessing, the dataset contains a total of 2,075,864 data samples. The dataset is partitioned into training, validation, and testing sets in a ratio of $8:1:1$. 

\begin{figure}[h]
    \centering
    \includegraphics[width=0.95\textwidth]{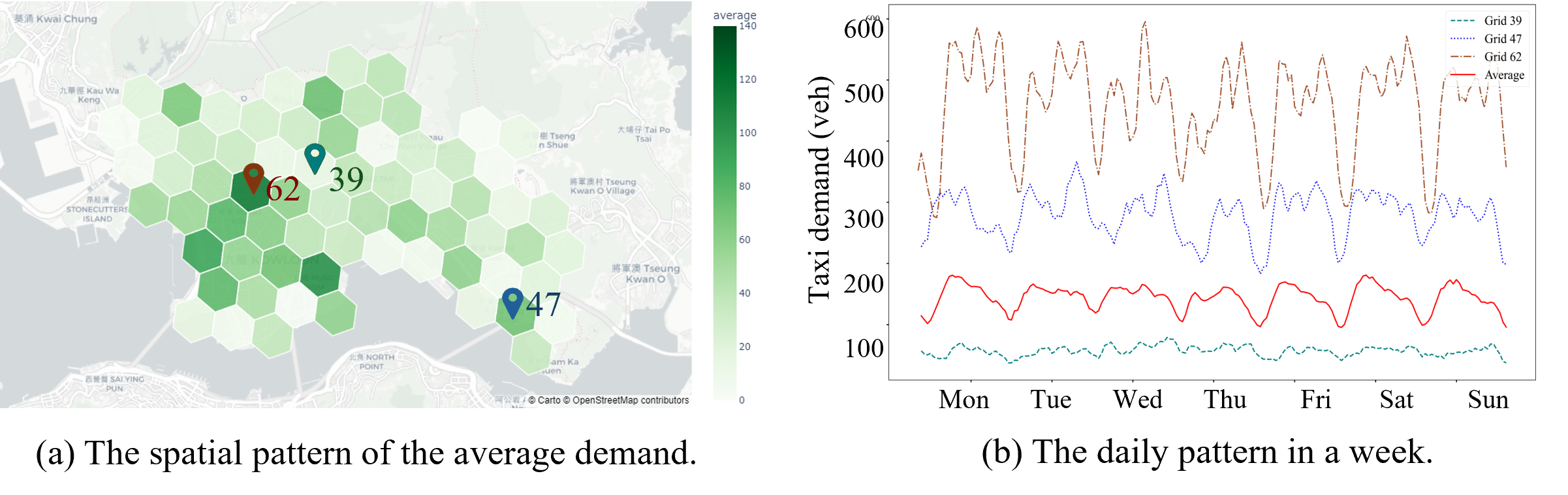}
    \caption{The spatio-temporal demand pattern in the research area.}
    \label{fig: demand pattern}
\end{figure}

Figure \ref{fig: demand pattern} visualizes the spatio-temporal taxi demand pattern in a week in the research area. From Figure \ref{fig: demand pattern} (a), we observe that high-demand areas are concentrated in the west-central and southern regions, particularly around Mong Kok and Hung Hom areas. Grid 62, located in the Mong Kok area, experiences the highest number of taxi requests. Figure \ref{fig: demand pattern} (b) illustrates the daily demand pattern in a week (Moving average with sliding window = 4) for three selected grids and on average. One can see that the temporal patterns vary across different grids with different demands. The average demand ranges from 100 to 200 vehicles per hour. Grid 62 exhibits the highest demand, ranging from 300 to 600 vehicles per hour, with three distinct demand peaks in the morning, noon, and late night. Grid 47 shows diverse peak hours in the morning, noon, and afternoon, occurring at around 8:00, 12:00, and 19:00. Meanwhile, the demand in grid 39 fluctuates smoothly at around 50 vehicles/hour.

We then compare the accuracy and computational efficiency performance of the proposed SPO framework with the baseline methods in mid and large-scale networks in dimensions 2,025 $(45\times 45)$ and 4,624 $(68\times 68)$, respectively. The total travel budget is set to 10,000. The TGCN module is utilized as the prediction module in both the SPO framework and baseline methods. The ratio of DVs to NDVs is set to $6\colon4$. 

\subsubsection{Training curves}

\begin{figure}[h]
    \centering
    \includegraphics[width=0.9\linewidth]{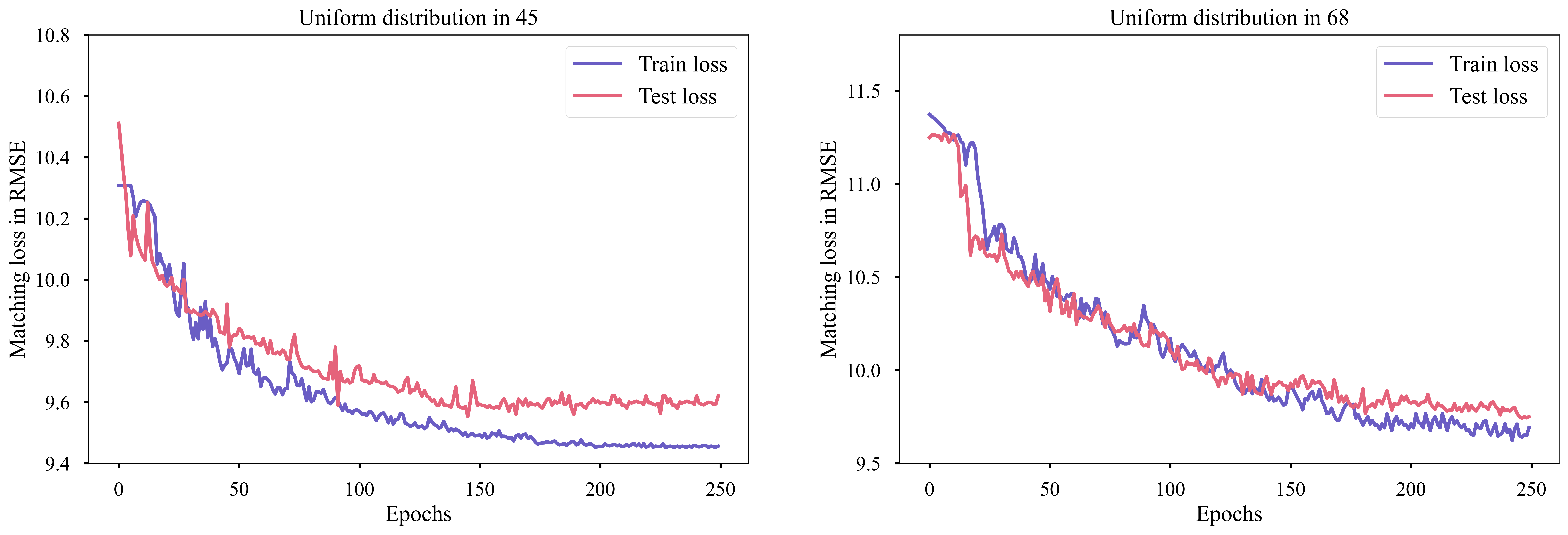}
    \caption{Training curves and test curves of SPO-A under the Uniform distribution in the mid-sized $(45\times 45)$ and large-scale $(68\times 68)$ network (Unit for RMSE: Vehicles/hour).}
    \label{fig: training curve}
\end{figure}

Figure \ref{fig: training curve} illustrates the training and testing curves of the proposed SPO-A framework under Uniform distribution in mid-sized and large-scale networks in case A. Both the training and testing curves for the large-scale $(68\times 68)$ dataset exhibit greater fluctuations compared to those of the mid-sized $(45\times 45)$  dataset, with both configurations converging around Epoch 250.

\subsubsection{Discussions}
This section analyzes the matching accuracy, computational efficiency, and detailed matching patterns comparing the PTO and proposed SPO frameworks.

\paragraph{Discussion on the matching accuracy}
The accuracy performance of the proposed SPO-A framework and other baseline methods is presented in Table \ref{tab: performance of the basic experiment}. 

\begin{table}[h]
    \caption{Performance of SPO-A and other baselines in HK dataset under Uniform target distribution (Unit for RMSE: Vehicles/hour).}
    \label{tab: performance of the basic experiment}
    \resizebox{\textwidth}{!}{
        \begin{tabular}{ccccccccccc}
            \toprule
            \multirow{2}{*}{\makecell[c]{Grid\\size}} 
            & \multicolumn{1}{c}{\multirow{2}{*}{Dimension}} 
            & \multicolumn{1}{c}{\multirow{2}{*}{\makecell[c]{Total\\budget}}} 
            & \multicolumn{2}{c}{\textbf{SPO-A(Ours)}} & \multicolumn{2}{c}{SPO-C} & \multicolumn{2}{c}{PTO} & \multicolumn{2}{c}{DON} \\ \cmidrule(l){4-11}
            & ~  & ~ & RMSE & SMAPE (\%) & RMSE & SMAPE (\%) & RMSE & SMAPE (\%) & RMSE & SMAPE (\%) \\ \midrule
            \multirow{4}{*}{$45 \times 45$} & \multicolumn{1}{c}{\multirow{4}{*}{2,025}}& \multicolumn{1}{c}{8000}& \textbf{9.072} & 31.623&9.510 & \textbf{31.551}&9.563 & 34.081 &14.517& 52.643 \\
            & \multicolumn{1}{c}{}& \multicolumn{1}{c}{10,000}& 9.647 & 31.552 & \textbf{9.541} & \textbf{31.457} & 9.551 & 33.987& 14.517 & 52.643\\
            & \multicolumn{1}{c}{}& \multicolumn{1}{c}{12000} & 9.589& 31.587& 9.593 & 31.558 & \textbf{9.554}& \textbf{31.523}& 14.517& 52.643 \\
            & \multicolumn{1}{c}{}& \multicolumn{1}{c}{15,000} & \textbf{9.059} & \textbf{31.426}& 9.507 & 31.431 & 9.563 & 34.087& 14.517& 52.643 \\\midrule
            \multirow{4}{*}{$68 \times 68$}  & \multicolumn{1}{c}{\multirow{4}{*}{4,624}}& \multicolumn{1}{c}{8,000}&\textbf{9.834} & \textbf{33.479} & 9.962 & 33.514 &10.701 & 38.166 &15.449&56.384\\
            & \multicolumn{1}{c}{}& \multicolumn{1}{c}{10,000}& \textbf{9.689} & \textbf{32.966}& 9.820 & 33.643 & 10.519& 36.955&15.449& 56.384 \\
            & \multicolumn{1}{c}{}& \multicolumn{1}{c}{12,000}& \textbf{9.291} & \textbf{30.504}& 9.678& 33.282 & 10.098 & 33.162& 15.449 & 56.384 \\
            & \multicolumn{1}{c}{} & \multicolumn{1}{c}{15,000} & \textbf{9.752}  & \textbf{33.277} & 10.096 & 32.478 & 10.547 & 37.495&  15.449  & 56.384 \\ \bottomrule
        \end{tabular}
        }
\end{table}

Under the mid-size network with $45\times45$ grids, the two end-to-end methods, SPO-A and SPO-C show comparable matching performance to the two-stage PTO method. Notably, the SPO-A method outperforms the PTO method under budget constraints of 8,000 and 15,000, showing an improvement of 5.20\%. However, in scenarios with a budget of 10,000 and 12,000, SPO-A slightly underperforms compared to SPO-C and PTO by less than 1.15\%. Under the large-scale network, one can see that the proposed method consistently outperforms PTO and SPO-C across all budget settings, achieving the highest average matching accuracy. On average, the SPO-A framework improved matching performance by 7.90\% compared to PTO and by 2.51\% compared to SPO-C across all budget settings. 

The results from different network scales indicate that the end-to-end framework presents an obvious enhancement over the two-stage methods under the predict-then-optimize paradigm. Moreover, the proposed SPO with alternating differentiation methods (SPO-A) method shows better and more robust performance in the large-scale network than in the mid-size network, validating the effectiveness of the alternating differentiation method in optimization. 

\paragraph{Discussion on the computational efficiency}

In this section, we evaluate the computational efficiency of the SPO-based frameworks, SPO-A and SPO-C, across both mid-size ($45 \times 45$) and large-scale ($68 \times 68$) networks, with a convergence threshold of $\xi = 5\times10^{-2}$. All experiments are conducted with a total budget of 10,000 under a Uniform distribution, and the other hyper-parameters follow the configuration specified in Table \ref{tab: hyperparameter}. Each experiment is repeated 5 times. The average total running time and the basic information of variables and constraints are reported in Table \ref{tab: computational efficiency}.

The results in Table \ref{tab: computational efficiency} indicate that the running time of both SPO-A and SPO-C increases when the dimension of variables increases from mid-size (2,025) to large-scale (4,624) networks. Notably, SPO-A exhibits a 4.01\% advantage over SPO-C in the large-scale setting, although SPO-A slightly under performs SPO-C for the mid-size network.

\begin{table}[h]
    \centering
    \caption{Comparison of the total running time of SPO-A and SPO-C with convergence threshold $\xi = 0.05 $ (Unit for time: $\times10^3$ sec) in Case A.}
    \addtolength{\tabcolsep}{10pt}
    \begin{tabular}{c|cc}
    \toprule
      & Mid-size & Large-scale \\
    \midrule
        Number of grids & 45 & 68 \\
        Number of variables & 2,025 & 4,624 \\\
        Number of constraints  & 6,120 & 13,940 \\
    \midrule
        Running time of SPO-A & 26.81 ($+$9.79\%) & 35.28 \textbf{($-$4.01\%)}\\
        Running time of SPO-C & 24.42 & 36.75 \\
    \bottomrule
    \end{tabular}
    \label{tab: computational efficiency}
\end{table}

\begin{figure}[h]
    \centering
    \includegraphics[width=0.6\linewidth]{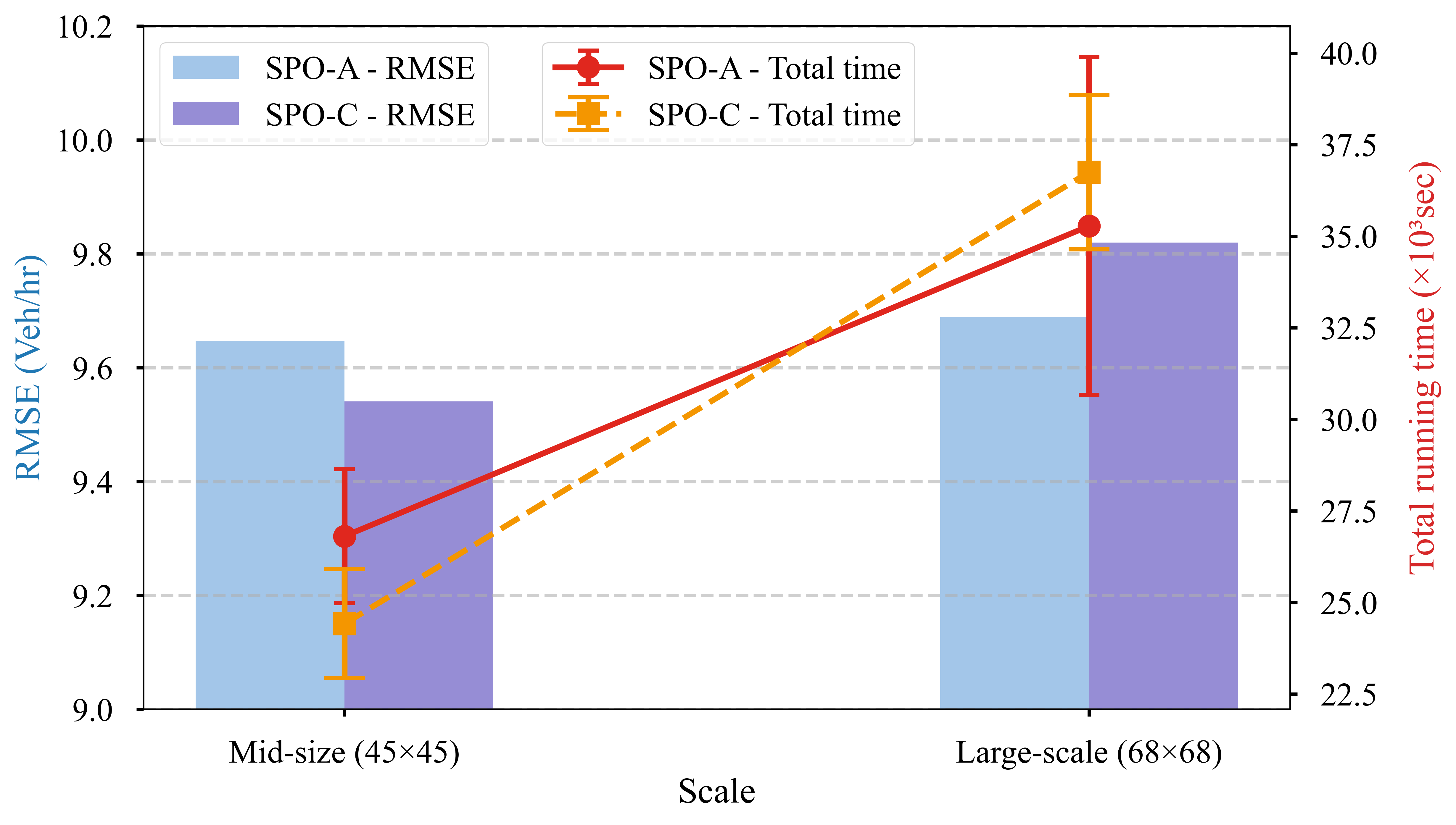}
    \caption{Comparison of matching accuracy and computational efficiency of SPO-A and SPO-C under mid-size and large-scale networks (Unit for accuracy: Vehicles/hour, unit for time: $\times10^3$ sec).}
    \label{fig: time vs accuracy}
\end{figure}

Figure \ref{fig: time vs accuracy} presents a comparative evaluation of the matching accuracy (measured in RMSE) and computational efficiency (total running time) for SPO-A and SPO-C under both mid-size and large-scale networks, with a convergence threshold $\xi = 5\times10^{-2}$. The results demonstrate that SPO-A achieves superior performance in the large-scale network, outperforming SPO-C in both accuracy and computational speed. This suggests that SPO-A is more scalable for high-dimensional problems, delivering competitive accuracy while reducing runtime. In the mid-size network, SPO-C exhibits marginally better performance, with SPO-A showing a modest 1.11\% decrease in RMSE matching accuracy and an approximate 9.79\% increase in running time. However, this difference in accuracy is relatively small, indicating that SPO-A remains a viable alternative even for mid-size applications.

\paragraph{Discussion on the prediction and matching patterns on the PTO and SPO frameworks}
\label{sec: discussion on prediction and matching patterns}
As introduced in Section \ref{sec: intro}, while the PTO framework may achieve better prediction accuracy in some cases, it can underperform in final decision accuracy. Table \ref{tab: example} provides a simple example of the relocation problems in VCS to illustrate how PTO fails in matching accuracy despite its predictive performance. Consider a scenario where the required number of vehicles in a grid is 100, but only 40 NDVs are available before relocating.

\begin{itemize}
    \item If the NDVs are overestimated to 44 in the prediction stage, the system will then plan to relocate 56 DVs to this area in optimization. In practice, the final allocation of DVS might reach 56. Then the total number of final vehicles after relocation would be 96. 
    \item If the NDVs are underestimated to 35 in the upstream prediction, the system will plan to relocate 65 DVs in optimization. However, achieving this higher requirement is more challenging, and the final allocation of DVs might only reach 62. The total number of final vehicles after relocation would be 102.
\end{itemize}

Overall, although overestimation of NDVs yields a smaller prediction deviation (overestimation: 4 vs. underestimation: 5), underestimation, despite its higher prediction error, can result in better final matching performance, as reflected in the lower matching deviation for DVs (overestimation: 4  vs. underestimation: 2). During peak hours, the tendency of PTO to overestimate demand, especially in high-demand areas (grids), often results in larger matching deviations. Because overestimation reduces the required number of relocation demands in the optimization stage, which will be easily fulfilled, inadvertently worsening the final matching discrepancy.

\begin{table}[h]
    \caption{Matching and prediction deviation comparison of the PTO in performing relocation tasks, with deviations measured in absolute terms. Unit for deviation: Vehicles/hour.}
    \centering
    \setlength{\tabcolsep}{2pt}
    \resizebox{\textwidth}{!}{
    \begin{tabular}{l|cc|ccc|c|cc}
    \toprule
        Case & \makecell{Total \\ required} & \makecell{Actual \\ NDVs} & \makecell{Predicted \\ NDVs} &\makecell{Required \\ DVs} &
        \makecell{Achieved \\ DVs} &
        \makecell{Final \\ vehicles} &
        \makecell{Predicted NDV \\ deviation} &
        \makecell{Matching DV \\ deviation} \\
    \midrule
        \textbf{Overestimation} & 100 & 40 & 44 & 56 & 56 & \textbf{96} & \textbf{4} & \textbf{4} \\
        Underestimation & 100 & 40 & 35 & 65 & 62 & 102 & 5 & 2 \\
    \bottomrule
    \end{tabular}
    }
    \label{tab: example}
\end{table} 

\begin{figure}[H]
    \centering
    \includegraphics[width=0.62\linewidth]{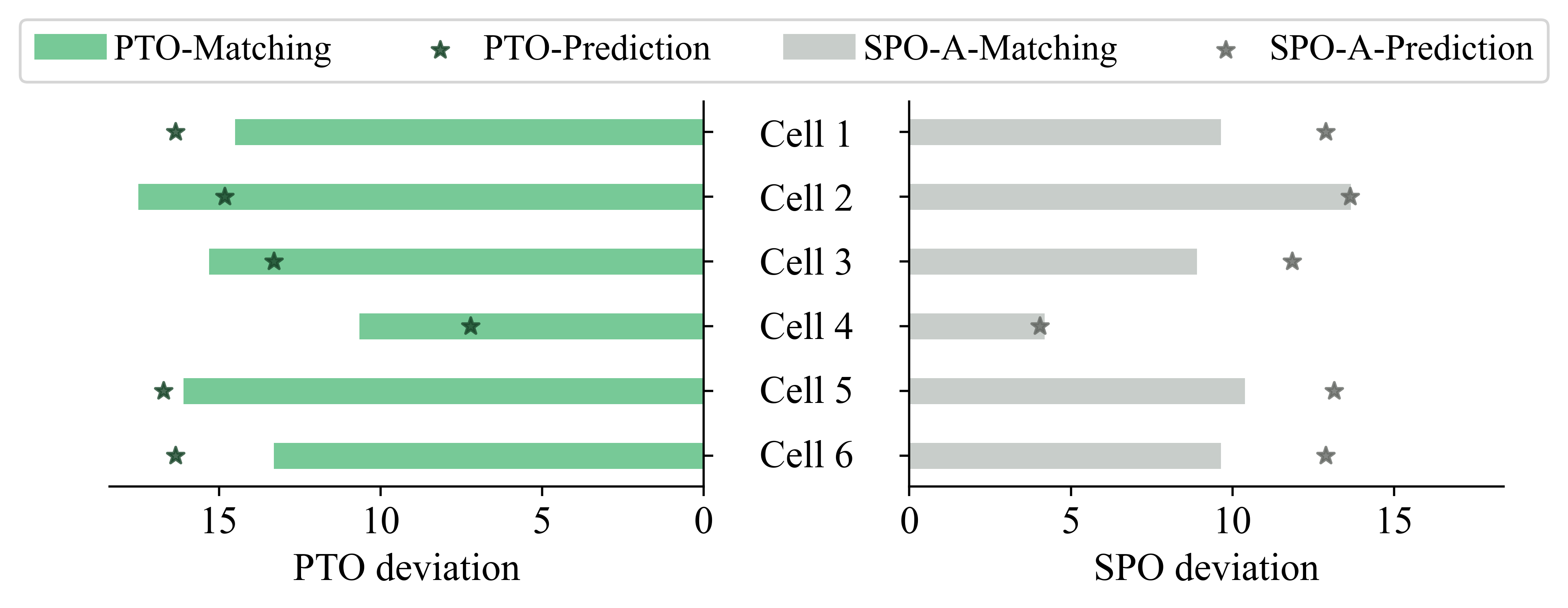}
    \caption{Matching and prediction deviation comparison of PTO and SPO-A frameworks under mid-size network in Case A. Unit for deviation: Vehicles/hour.}
    \label{fig: counterexample}
\end{figure}

We further investigate this phenomenon in the experiments for Case A, where we compare the prediction and matching deviations of the PTO and SPO-A frameworks across six selected grids at 10:00 AM under a mid-size network with a Uniform distribution, as shown in Figure \ref{fig: counterexample}. While both frameworks exhibit similar prediction accuracy across all 45 grids (PTO RMSE: 8.593 vs. SPO-A RMSE: 8.625), the SPO-A framework demonstrates superior overall matching performance (PTO: 9.551 vs. SPO-A: 9.647). However, in the high-demand six grids illustrated in Figure \ref{fig: counterexample}, PTO exhibits substantial prediction overestimation and larger matching deviation compared to SPO-A. This suggests that while the PTO framework demonstrates slightly better overall prediction performance in terms of RMSE (weighted across all grids), its tendency to overestimate in critical high-demand areas results in compounded errors in matching deviation, ultimately degrading the final matching accuracy. In contrast, the SPO-A framework, which is trained by directly minimizing task-specific matching errors with the assistance of prediction errors, corrects both intermediate predictions and final matching deviations more effectively than the PTO framework and therefore shows better overall matching accuracy.

\begin{figure}[h]
    \centering
    \includegraphics[width=0.99\linewidth]{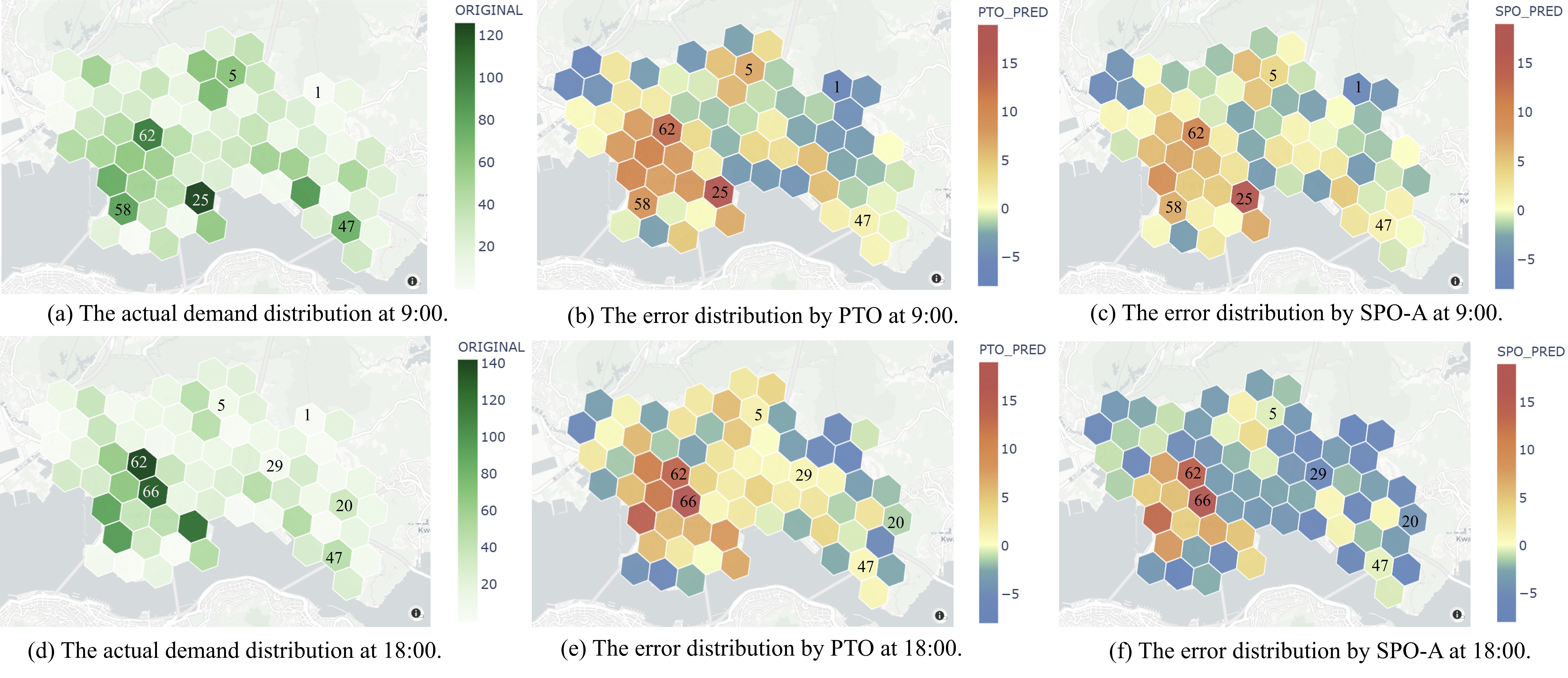}
    \caption{The spatial error distribution of predicted demand at 9:00 and 18:00 by the PTO and the proposed SPO-A framework under large-scale (68$\times$68) networks. In (a) and (d), the underlying green represents the actual demand, with darker shades indicating higher values. In (b-c) and (e-f), the prediction error is defined as the deviation between predicted and actual demand, which is overlaid in color: red for overestimation, blue for underestimation, and yellow for near-zero errors.}
    \label{fig: error distribution}
\end{figure}

We then analyze the spatio-temporal patterns of demand prediction errors for the PTO and proposed SPO-A frameworks on the large-scale (68×68) network during the morning (9:00) and afternoon (18:00) peak hours on Thursday for Case A, as illustrated in Figure \ref{fig: error distribution}. Generally, the afternoon peak exhibits higher total demand and greater regional variation than the morning peak. SPO-A achieves more balanced predictions with lower bias during the lower-demand morning peak (Prediction RMSE: SPO-A: 8.782 vs. PTO: 9.175). However, during the high-variation afternoon peak, SPO-A shows a tendency toward underestimation, resulting in a higher RMSE than PTO (SPO-A: 9.583 vs. PTO: 9.272) but shows higher matching accuracy (SPO-A: 9.792 vs. PTO: 10.012 at 18:00). For a detailed spatial analysis, we select six representative grids in each time interval. In high-demand downtown areas, both frameworks tend to overestimate demand (e.g., cells 25 and 62 at 9:00; cell 66 at 18:00). But the PTO framework tends to show larger over-estimation than SPO-A. In transition areas surrounding high-demand downtown regions (e.g., cells 5, 47), prediction errors for both frameworks are generally low, fluctuating near zero. The degree of this overestimation is consistently lower for SPO-A. In peripheral areas, demand is generally underestimated, and here, SPO-A exhibits a relatively larger underestimation bias than PTO (e.g., cells 1, 20, and 29). To conclude, although the relatively large under-estimation of the SPO-A leads to final larger prediction errors, the reduced over-estimation in high-demand areas compensates and primarily contributes to the final overall superior matching performance. In terms of PTO, the pronounced overestimation in high-demand and surrounding transition areas during peak hours adversely impacts the final matching accuracy. This finding is consistent with analysis in mid-size networks.

\subsubsection{Ablation study}

In this section, we present four groups of ablation models with three different prediction models (TGCN, DCRNN \citep{li2018dcrnn_traffic}, and ASTGCN \citep{guo2019attention}) to justify the effect of different prediction and optimization modules: SPO-A, SPO-C, PTO, NOP (no prediction but with the optimization), and DON (no prediction and no optimization). The four grouped models are examined in both mid and large-scale networks under Uniform and Gaussian target sensing distributions. The ratio of DVs to NDVs is $6\colon4$, and the total budget is 10,000.

\paragraph{Matching performance}
\label{sec: matching accuracy discussion}
We first discuss the matching accuracy of different prediction models. 

The matching results are summarized in Table \ref{tab: ablation}. Overall, the SPO-A methods with different prediction modules achieve robust and satisfactory performance for all scale network sizes and all target distributions in matching accuracy. 

\begin{table}[h]
    \caption{The comparison of the matching performance of prediction modules in different frameworks under different networks and target distributions (Unit for RMSE: Vehicles/hour).}
    \label{tab: ablation}
    \centering
    \resizebox{\textwidth}{!}{
        \begin{tabular}{cccccccccc}
            \toprule
            \multirow{2}{*}{\makecell[c]{Prediction\\module}} & \multirow{2}{*}{Framework} & \multicolumn{2}{c}{45-Uniform} & \multicolumn{2}{c}{45-Gaussian} & \multicolumn{2}{c}{68-Uniform} & \multicolumn{2}{c}{68-Gaussian} \\
            \cmidrule(l){3-10}
                ~   & ~    & RMSE & SMAPE (\%) & RMSE & SMAPE (\%) & RMSE  & SMAPE (\%)  & RMSE & SMAPE (\%)  \\
                \midrule
                \multirow{5}{*}{TGCN} & \textbf{SPO-A (Ours)}& 9.647&31.552&\textbf{9.692}&\textbf{36.673}&\textbf{9.689}&\textbf{32.966}&\textbf{10.312}&\textbf{44.287} \\
                ~ & SPO-C& \textbf{9.541}&\textbf{31.457}&9.858&36.680&9.820&33.643&10.642&46.029 \\
                ~ & PTO & 9.551&33.987&9.704&36.742&10.519&36.955&11.527&48.112 \\
                ~ & NOP & 13.465&47.988&13.581&53.551&14.569&54.450&15.722&52.675 \\
                ~ & DON & 14.517&52.643&19.352&75.164&15.449&56.384&20.583&76.103 \\
                \midrule
                \multirow{5}{*}{DCRNN} & \textbf{SPO-A (Ours)}& 8.779&30.447&\textbf{8.126}&\textbf{36.368}&\textbf{9.223}&\textbf{35.507}&\textbf{10.186}&\textbf{42.956}\\
                ~ & SPO-C& \textbf{8.747}&\textbf{26.317}&8.327&36.602&9.710&36.583&10.480&43.236 \\
                ~ & PTO & 8.781&28.895&8.804&40.760&9.710&36.853&11.957&51.171 \\
                ~ & NOP & 13.465&47.988&13.581&53.551&14.569&54.450&15.722&52.675 \\
                ~ & DON & 14.517&52.643&19.352&75.164&15.449&56.384&20.583&76.103 \\
                \midrule
                \multirow{5}{*}{ASTGCN} & \textbf{SPO-A (Ours)}& \textbf{8.476}&\textbf{29.591}&\textbf{9.430}&\textbf{46.292}&\textbf{9.225}&\textbf{30.316}&\textbf{10.667}&\textbf{44.606}\\
                ~ & SPO-C& 8.486&29.980&9.496&36.479&9.644&36.360&10.669&44.702 \\
                ~ & PTO & 9.694&30.301&9.986&47.338&9.721&32.610&11.071&45.366 \\
                ~ & NOP & 13.465&47.988&13.581&53.551&14.569&54.450&15.722&52.675 \\
                ~ & DON & 14.517&52.643&19.352&75.164&15.449&56.384&20.583&76.103 \\
            \toprule
        \end{tabular}
    }
\end{table}

\begin{enumerate}[{(1)}]
    \item \textbf{Justifying the matching performance of different prediction models}: When comparing the SPO-A methods with two-stage methods incorporating three similar prediction modules—TGCN, DCRNN, and ASTGCN, we observe that methods utilizing the DCRNN prediction module consistently demonstrate superior matching performance relative to the other two prediction modules. Furthermore, the SPO-A method outperforms other benchmark models in all scenarios except for the Uniform distribution with 45 grids. This indicates the robustness of the SPO framework across various network sizes and target distribution scenarios.
    
    \item \textbf{Justifying the use of the prediction module}: Comparing NOP with SPO-A, we observe a significant decrease in matching performance by 34.01\% on average when prediction is removed while retaining optimization. This indicates the crucial role of the prediction module in achieving higher matching performance. 
    
    \item \textbf{Justifying the use of the alternating differentiation method in optimization}: Comparing SPO-A and SPO-C, it is evident that substituting the alternating differentiation method with the implicit differentiation method results in a significant decline in matching accuracy in large-scale networks, despite only minor differences between the two frameworks in mid-size networks. This highlights the importance of the alternating differentiation method in large-scale networks.

\end{enumerate}

\paragraph{Prediction performance}

\begin{figure}[h]
    \centering
    \includegraphics[width=0.8\linewidth]{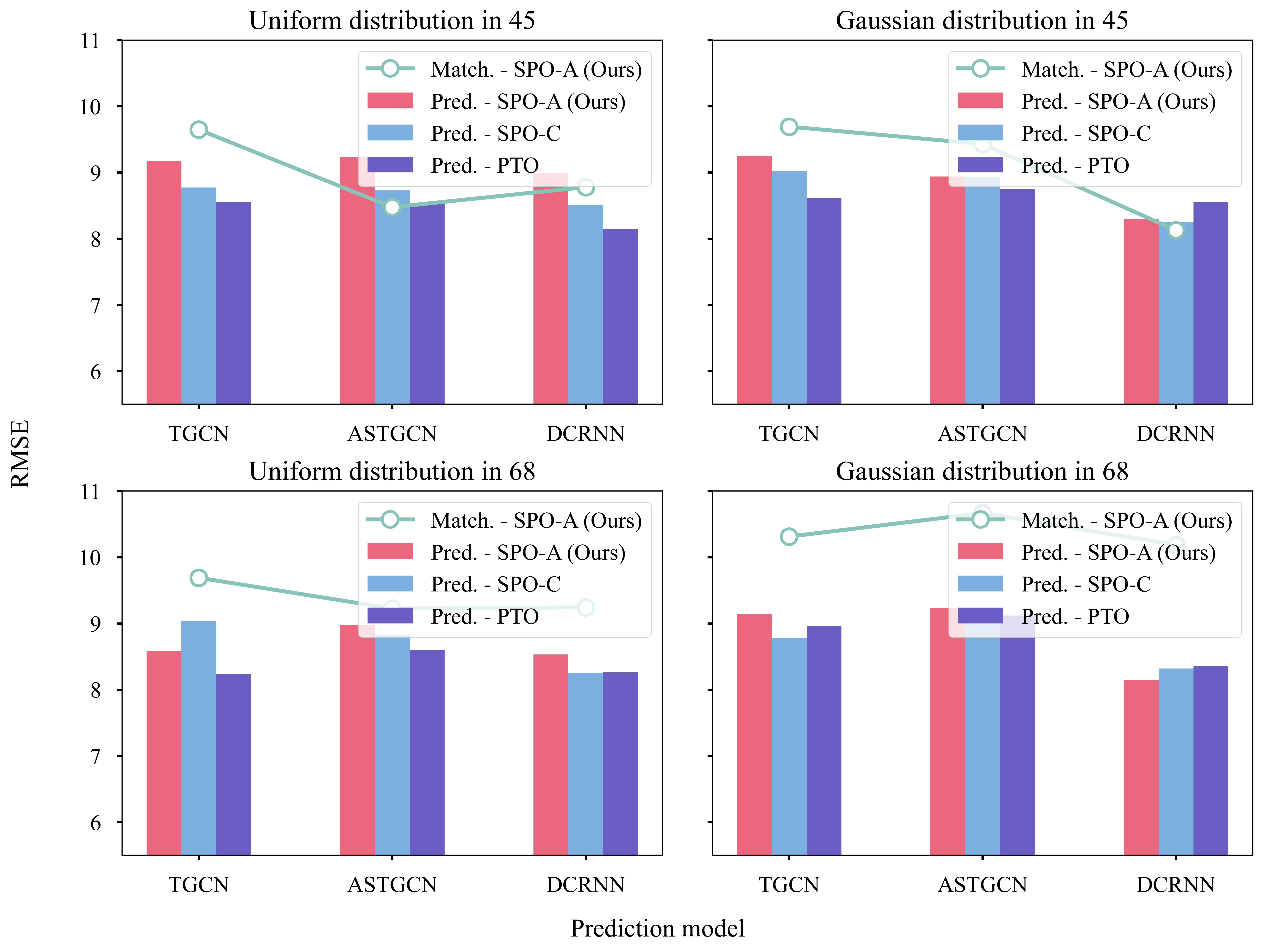}
    \caption{The prediction accuracy of different prediction models in SPO-A, SPO-C, and PTO framework and the matching accuracy of SPO-A framework under Uniform and Gaussian distribution in mid-size and large-scale networks (Unit for RMSE: Vehicles/hour; Pred., prediction; Match., matching).}
    \label{fig: prediction}
\end{figure}

We then proceed to compare the prediction accuracy of different prediction modules, including TGCN, DCRNN, and ASTGCN in SPO-C, SPO-A, and PTO framework, and compare the prediction accuracy with matching accuracy under Uniform and Gaussian distribution in mid-size and large-scale networks, as is illustrated in Figure \ref{fig: prediction}. The implementation details of the hyperparameters of each prediction module are provided in Appendix \ref{apd: hyperparameter}. 

\begin{enumerate}[{(1)}] 
        \item \textbf{Justifying the prediction accuracy across different frameworks}. For the SPO-A, SPO-C, and PTO frameworks with various prediction models, it is noteworthy that the prediction accuracy of the SPO framework (including both SPO-A and SPO-C) is slightly lower (by 0.2-0.6 in RMSE) than that of the two-step PTO framework in some cases (in off-peak hours), despite the SPO framework achieving better final matching accuracy on average. This difference can be attributed to the SPO framework's training process, which involves a combined prediction and matching divergence, whereas the PTO framework is trained solely based on prediction divergence. This suggests that incorporating matching divergence can reduce the final matching error but may slightly increase the mid-term prediction error. 
    
    \item \textbf{Justifying the prediction accuracy of different prediction models}. DCRNN demonstrates the highest prediction accuracy, outperforming TGCN and ASTGCN, with ASTGCN being the least accurate. This trend is also consistent with the matching performance of the three prediction models discussed in Section \ref{sec: matching accuracy discussion}.

    \item \textbf{Comparing prediction accuracy with matching accuracy}. In most cases, the matching RMSE is higher than the prediction RMSE. However, in the Uniform and Gaussian distributions within mid-size networks with 45 grids, we observed that the final matching RMSE is lower than the mid-term prediction RMSE. This is because matching accuracy depends on both prediction accuracy and relocation accuracy, and the relocation RMSE can be relatively lower based on these specific prediction results.

\end{enumerate}

\subsubsection{Sensitivity analysis}

This section presents four sensitivity analyses conducted on the proposed SPO methods and baselines, focusing on the convergence threshold, target sensing distributions, control ratios, and the penalty term. The first sensitivity analysis assesses how runtime scales with the convergence threshold under varying precision settings. The second sensitivity analysis on potential target distributions examines the robustness and applicability of the proposed SPO framework across diverse sensing tasks in VCS by evaluating its performance under varying target distributions. The third analysis investigates the influence of the control ratio, specifically the impact of NDVs and DVs in different proportions, on the execution of sensing tasks. In the fourth sensitivity analysis, cross-validation is used to select the appropriate value for the penalty term hyper-parameter. These analyses together provide insights into the robustness, adaptability, and effectiveness of the proposed framework under different realistic operational conditions in VCS.

\paragraph{Sensitivity analysis on the convergence threshold}

\begin{figure}[h]
    \centering
    \includegraphics[width=0.9\linewidth]{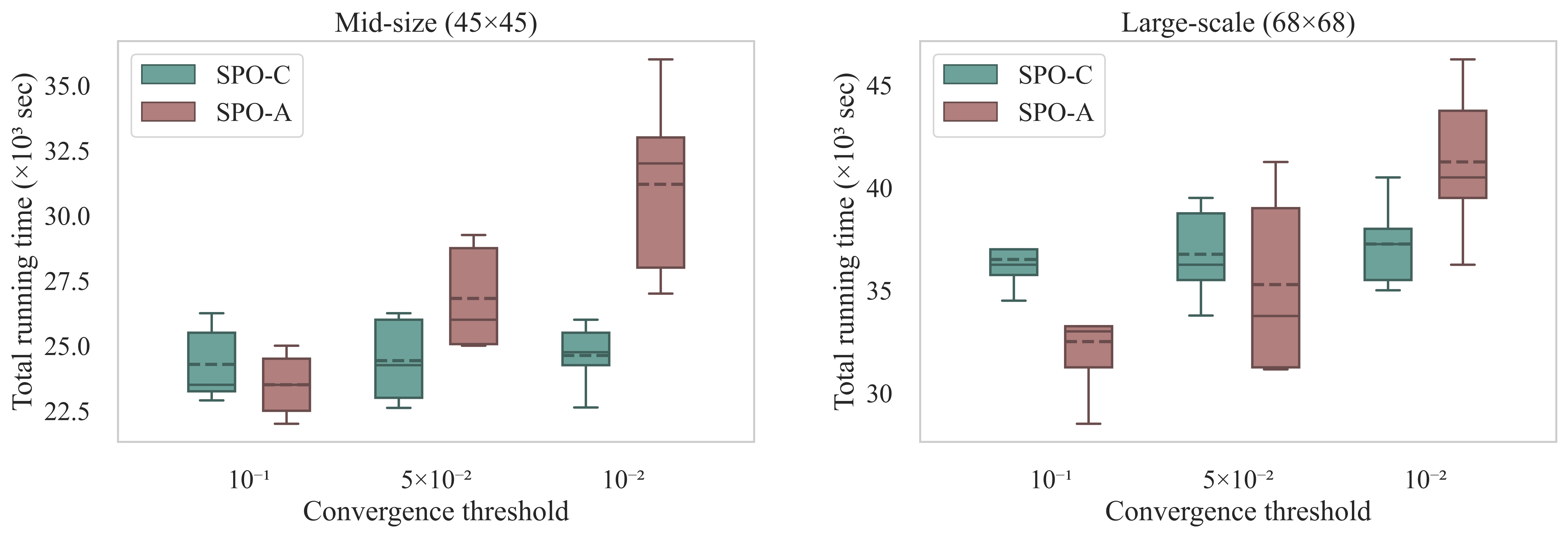}
    \caption{Comparison of the computational efficiency of SPO-A and SPO-C framework on the mid-size and large-scale networks on varying convergence thresholds. (Unit for running time: $\times10^3$ sec).}
    \label{fig: sensitivity_efficiency}
\end{figure}

This section systematically evaluates the relationship between computational efficiency and accuracy performance within the SPO-based framework, SPO-A and SPO-C. we conduct controlled experiments comparing SPO-A and SPO-C across mid-size $(45\times45)$ and large-scale $(68\times 68)$ networks, with the threshold $\xi \in \{10^{-1}, 5 \times 10^{-2}, 10^{-2}\}$. All experiments are conducted with a total budget of 10,000 under a Uniform distribution while maintaining consistent hyperparameters as specified in Table \ref{tab: hyperparameter}. TGCN serves as our prediction module. Each configuration is replicated 5 times. Table \ref{tab: efficiency_SPO_A} evaluates the computational efficiency and accuracy performance of the SPO-A framework across mid-size and large-scale networks with varying convergence thresholds, and Figure \ref{fig: sensitivity_efficiency} compares the computational efficiency of both SPO-A and SPO-C frameworks across mid-size and large-scale networks with different convergence thresholds.

\begin{table}[h]
    \centering
    \caption{Comparison of matching accuracy and computational efficiency of the SPO-A framework under mid-size and large-scale networks with varying convergence thresholds.}
    \label{tab: efficiency_SPO_A}
    \begin{tabular}{c|ccccc}
        \toprule
         \multirow{2}{*}{ Threshold $\xi$}  & \multicolumn{2}{c}{Mid-size ($45 \times 45$)} & \multicolumn{2}{c}{Large-scale ($68 \times 68$)} \\
        \cmidrule(lr){2-3} \cmidrule(lr){4-5}
        ~ & Accuracy (RMSE) & Efficiency (Time) & Accuracy (RMSE) & Efficiency (Time) \\
        \midrule
        $10^{-1}$ & 9.995 & 23.52 & 10.137 & 32.54 \\
        $5\times10^{-2}$ & 9.647 & 26.81 & 9.689 & 35.28 \\
        $10^{-2}$ & 9.501 & 31.20 & 9.465 & 41.25 \\
        \bottomrule
\end{tabular}

\smallskip
\footnotesize \textit{Note:} Accuracy measured in RMSE (Vehicles/hour), efficiency in total running time ($\times 10^3$ sec).
\end{table}

Combined the results in Table \ref{tab: efficiency_SPO_A} and Figure \ref{fig: sensitivity_efficiency}, we summarize four key findings:

\begin{itemize}
    \item \textit{Accuracy improvement}:  A consistent improvement in matching accuracy (measured in RMSE) of SPO-A is observed with smaller threshold values across both mid-size and large-scale networks. But this accuracy gain comes at a substantial computational cost, with the total running time increasing approximately 21.11\% to 24.62\% when the threshold $\xi$ tightens from $10^{-1}$ to $10^{-2}$. 
    \item \textit{Scalability advantage}: SPO-A exhibits superior computational efficiency in large-scale networks compared to SPO-C, especially at $\xi = 10^{-1}$ and $5\times10^{-2}$. This advantage diminishes in mid-size networks, where SPO-C shows marginally better performance. 
    \item \textit{Threshold sensitivity}: SPO-A displays greater sensitivity to threshold precision in computational efficiency, evidenced by larger runtime standard deviations (maximum 16.17) compared to SPO-C (maximum 9.42) across all thresholds and all network scales.
    \item \textit{Accuracy-Efficiency trade-off}: In large-scale networks, SPO-A achieves higher matching accuracy by 6.62\% as $\xi$ tightens from $ 10^{-1}$ to $5 \times 10^{-2}$, although requires approximately 26.76\% additional computation time.
\end{itemize}

Overall, SPO-A demonstrates better scalability than SPO-C in high-dimensional settings, achieving comparable accuracy with relatively lower running time. Besides, selecting an appropriate convergence threshold for SPO-A is critical to maintaining both acceptable accuracy and stable computational efficiency across varying operational conditions. Our experiment results identify $5\times 10^{-2}$ as the optimal convergence threshold for the SPO-A framework in large-scale networks, achieving a balanced accuracy gain and comparable running time. These findings also highlight the context-dependent nature of threshold selection, where strict thresholds (smaller $\xi$) favor matching accuracy while moderate thresholds (larger $\xi$) prioritize computational efficiency.

\paragraph{Sensitivity analysis on the uncertainty of target sensing distributions}
\label{subsec: uncertainty of distribution}

\begin{figure}[h]
    \centering
    \includegraphics[width = 0.98\textwidth]{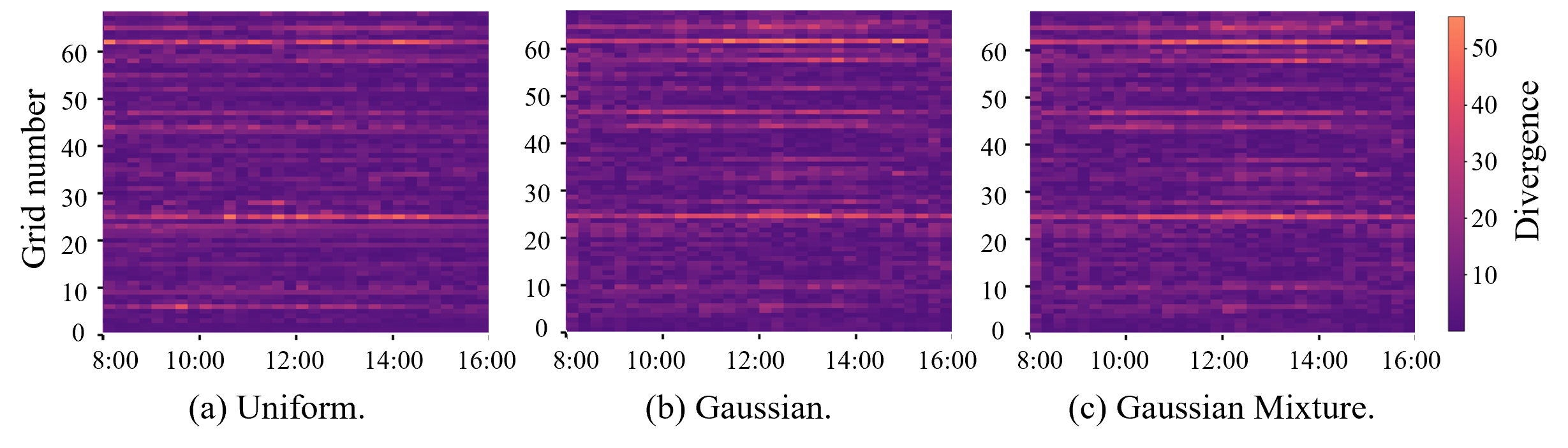}
    \caption{Sensitivity analysis of the SPO framework on different time intervals under Uniform, Gaussian, and Gaussian Mixture distributions in large-scale networks.}
    \label{fig: density plot}
\end{figure}

To evaluate the robustness of the framework under varying target distributions, we conduct a sensitivity analysis using Uniform, Gaussian, and Gaussian Mixture distributions. This analysis is performed at 15-minute intervals over a 24-hour period to comprehensively assess the adaptability of the proposed framework to different sensing scenarios. Note that the Gaussian Mixture distribution diverges most from the original distribution. The ratio of DVs to NDVs is set to $6\colon4$, and the total budget is set to 10,000. Figure \ref{fig: density plot} shows the absolute divergence in each grid from 8:00 to 16:00. Each pixel in each sub-figure represents the absolute value of the divergence between the target distribution and the matching distribution. The lighter the color is, the larger the divergence is. 

From Figure \ref{fig: density plot}, it is apparent that Grid 62 and Grid 25 have larger divergence, especially from 10:00 to 12:00 in all three target distributions. This discrepancy may be attributed to the anticipation of the prediction from free vehicle movements based on weekly patterns. Grid 62 in the Mong Kok area and Grid 25 in the Hunghom area experience a surge in demand during noon. When the distribution of NDVs deviates from the target distribution, the remaining DVs struggle to compensate for this divergence. Conversely, in grids with smoother demand fluctuations, the matching divergence is less. Overall, significant fluctuations in actual taxi demand may challenge the ability of the SPO-A framework to effectively satisfy matching distributions. However, if the target distribution diverges substantially from the original distribution, SPO-A is likely to outperform other baseline methods.

\paragraph{Sensitivity analysis on the uncertainty of control ratios}

\begin{figure}[h]
    \centering
    \includegraphics[width = 0.98\textwidth]{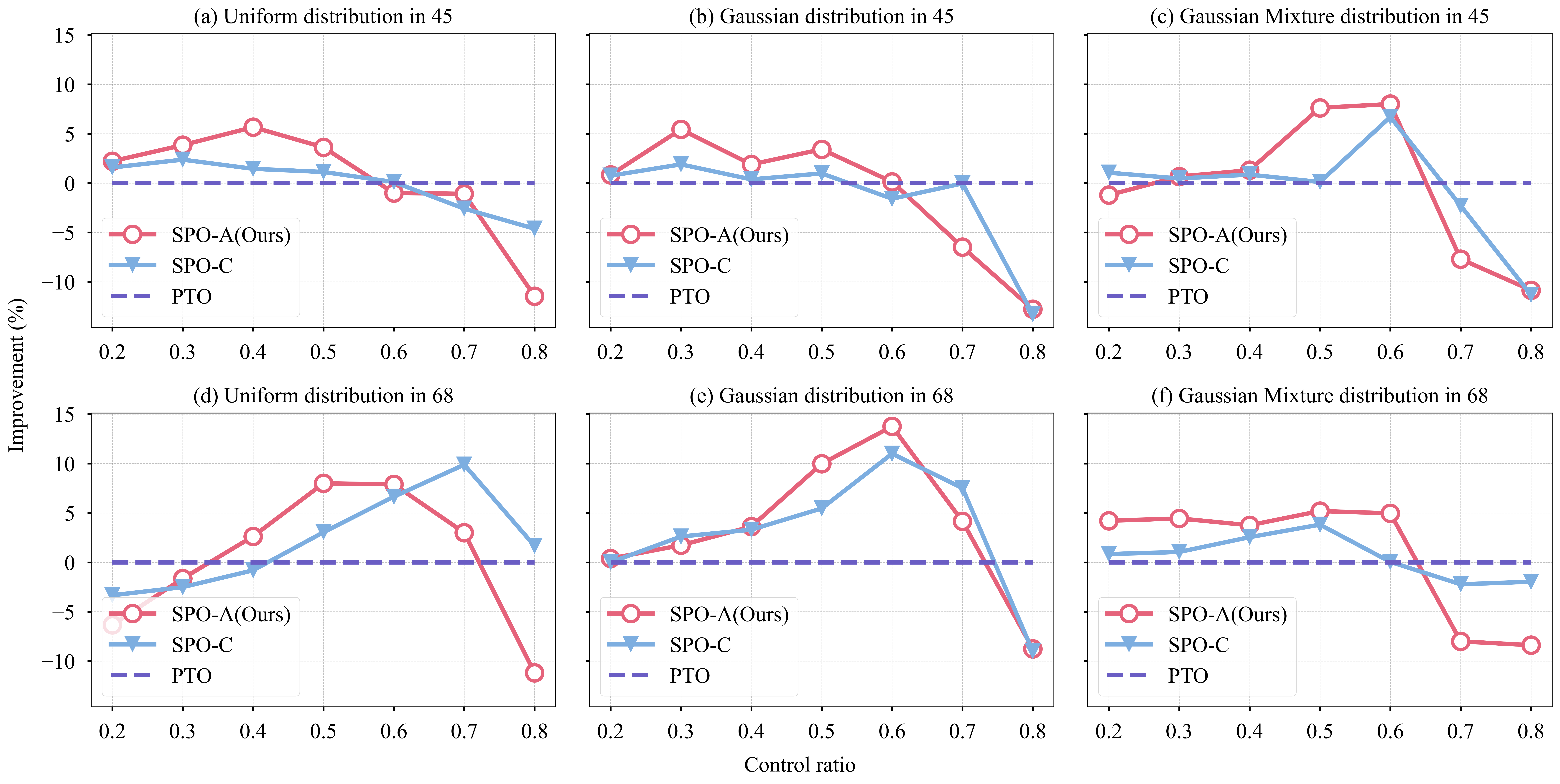}
    \caption{The improvement in the matching performance of SPO-A with SPO-C and PTO frameworks under Uniform, Gaussian, and Gaussian Mixture distributions in mid-size and large-scale networks, using the matching RMSE of the PTO framework as a reference.}
    \label{fig: line plot}
\end{figure}

In this section, we analyze how matching performance varies among the SPO-A, SPO-C, and the PTO framework across different control ratios in real traffic networks, as illustrated in Figure \ref{fig: line plot}. As we assume that the DVs will fully obey the assignment from the dispatching center, we implement a straightforward classification approach: any DV rejecting a task is classified as an NDV. Therefore, to account for varying compliance levels, we perform sensitivity analysis across a range of DV-to-NDV control ratios (20\%-80\%), with increments of 10\%, in both mid and large-scale networks under Uniform, Gaussian, and Gaussian Mixture distributions to simulate different operational scenarios. We compare the improvement in matching accuracy of the proposed SPO-A framework against the SPO-C and PTO frameworks, using the RMSE of the PTO framework as a reference.

In the mid-size network, as shown in Figure \ref{fig: line plot} (a), (b), and (c), with control ratios ranging from 20\% to 60\%, the SPO-A framework consistently outperforms the SPO-C and PTO frameworks in most cases. The highest improvement of the SPO-A, 7.52\%, is observed at a 60\% control ratio under the Gaussian Mixture distribution. Additionally, both the SPO-A and SPO-C frameworks demonstrate better matching performance than the PTO framework within this control ratio range. The SPO-C framework performs comparably to the PTO method, showing an improvement of around 2\% over the PTO framework. 

In the large-scale network, as shown in Figure \ref{fig: line plot} (d), (e), and (f), the SPO-A framework achieves even better matching performance compared to the mid-size network, with a peak improvement of 13.75\% under the Gaussian distribution compared with the PTO framework. The SPO-A method also significantly outperforms both the SPO-C and PTO frameworks, showing more substantial improvements over these two baselines.

However, with control ratios exceeding 60\% in mid-size and 70\% in large-scale networks, the matching performance of the SPO-A framework declines, falling below that of the SPO-C and PTO methods. This decline is attributed to precision errors occurring during the convergence process in the alternating differentiation iterations.

Overall, the SPO-A framework demonstrates significantly better performance than the SPO-C and PTO frameworks within the control ratio range of 0.3 to 0.7. As the control ratio increases, the precision errors in the alternating differentiation method can enlarge, negatively impacting the final matching performance.

\begin{figure}[h]
    \centering
    \includegraphics[width=0.5\linewidth]{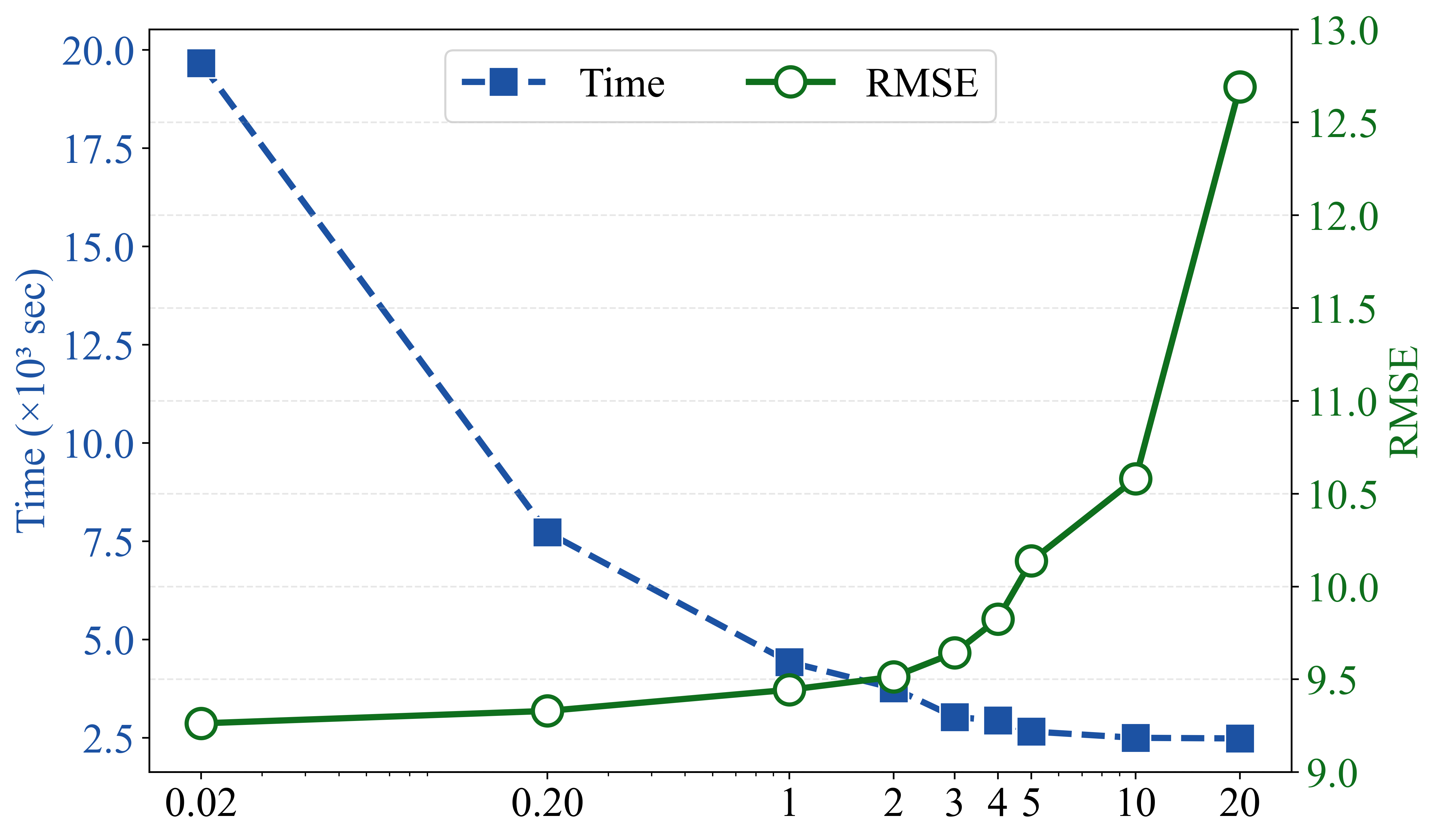}
    \caption{Sensitivity analysis on the penalty term $\rho$ in the SPO-A framework (in log scale).}
    \label{fig: sensitivity penalty term}
\end{figure}

\paragraph{Sensitivity analysis on the penalty term in SPO-A}
\label{sec: penalty term}
The penalty term $\rho$ plays a crucial role in balancing computational efficiency and accuracy within the SPO-A framework. It acts as an adaptive learning rate for dual variable updates while simultaneously transforming the original hard constraints into a more flexible soft solution space. A large $\rho$ can accelerate convergence speed, show faster initial convergence. However, a large $\rho$ may lead to overfitting, cause oscillations near the optimum, and even end in a local optimum. Conversely, a small $\rho$ slows down convergence, provides smoother updates, but increases the total training time, especially in large-scale problems. In this section, we perform cross-validation on the penalty term $\rho$ in the SPO-A framework in the HK dataset under size $45 \times 45$  to identify the optimal value that balances accuracy and computational efficiency. We evaluate $\rho$ over the set $\{0.02, 0.2, 1, 2, 3, 4, 5, 10, 20\}$ with the result presented in Figure \ref{fig: sensitivity penalty term}. It is observed that as $\rho$ increases, the RMSE rises significantly beyond $\rho =10$, while the training time decreases marginally. The efficiency and accuracy curves intersect at $\rho = 2$, indicating that $\rho = 2$ strikes a reasonable balance between accuracy and efficiency for the HK dataset. Although the RMSE continues to decrease with a minimum value of $\rho = 0.02$, the training time increases by up to four times compared to $\rho = 2$. Within the range of $\rho = 1$ to $\rho = 5$, the RMSE and efficiency fluctuate slightly. These results suggest that, in selecting an optimal penalty term $\rho$, it is advisable to first evaluate across different orders of magnitude, followed by fine-tuning within the chosen range to achieve the best trade-off between accuracy and computational efficiency.

\subsection{Case B: Large-scale ride-hailing dataset}

To further validate the adaptability of the proposed SPO framework in real-world scenarios, we conduct an additional experiment using a larger dataset from central Chengdu, China.

\subsubsection{Research area}

\begin{figure}[h]
    \centering
    \includegraphics[width=0.95\linewidth]{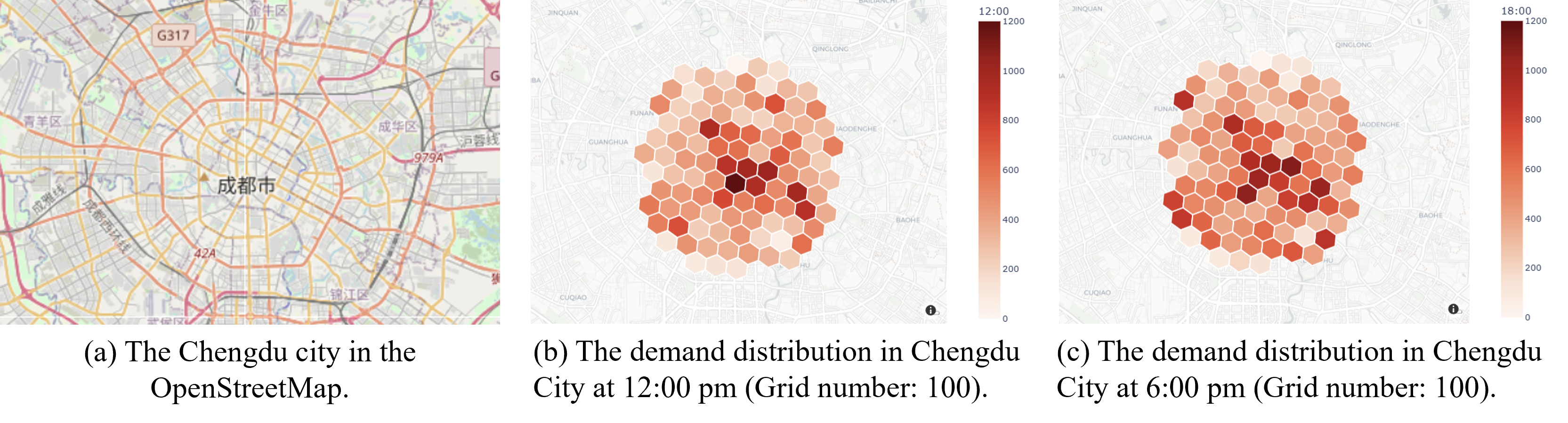}
    \caption{Overview of the research area in Chengdu City, China. Grids with darker shading represent higher taxi demand.}
    \label{fig: overview cdc}
\end{figure}

The dataset comprises two weeks of taxi trajectory data between August 3 and August 16, 2014,  sourced from the Open Dataset, with demand aggregated at 15-minute intervals. After preprocessing, the dataset contains 9,985,238 samples, approximately 5 times the total volume of the Hong Kong dataset (Case A). Figure \ref{fig: overview cdc}(a) presents the study area mapped on OpenStreetMap. Figures \ref{fig: overview cdc}(b) and (c) illustrate the spatial distribution of travel demand across 100 grids (H8 resolution) at 12:00 and 18:00, respectively. The visualization reveals that high-demand grids are predominantly concentrated in central areas, with demand intensity decreasing radially outward. The comparison reveals significantly higher demand during the evening peak (18:00) compared to noon (12:00), as clearly shown by the increased density of darker-shaded grids in Figures \ref{fig: overview cdc}(c). 

\subsubsection{Experimental configuration}
The prediction module in Case B extends beyond Case A by incorporating three distinct feature categories: 1) historical taxi demand per grid, 2) weather conditions, and 3) POI distributions. For weather data, we utilize hourly temperature (continuous) and precipitation (categorical) measurements obtained from Weather Spark \citep{weatherspark}. The POI feature encompasses 13 facility categories within each grid, including transportation hubs, tourist attractions, public amenities, dining establishments, and shopping centers.

To ensure comprehensive evaluation, we examine three network scales ($60\times60$, $80\times80$, and $100\times100$ grids) representing progressively larger urban areas. Vehicle compliance is assessed through systematic variation of the DV-to-NDV control ratio from 20\% to 80\% in 10\% increments across all scenarios. Furthermore, we employ three distinct target sensing distributions (Uniform, Gaussian, and Gaussian Mixture) to prevent model overfitting to any single demand pattern. Complete hyperparameter configurations are detailed in Table \ref{tab: hyperparameter}, while the dynamic weighting strategy specifications are provided in Table \ref{tab: weight setting}.

\subsubsection{Accuracy and efficiency performance}
We evaluate the matching accuracy of the proposed SPO-A framework against SPO-C and PTO across three large-scale network configurations in Case B, considering Uniform, Gaussian, and Gaussian Mixture distributions. As shown in Figure \ref{fig: lineplot cdc}, the results demonstrate three key findings consistent with Case A: 1) Both SPO-A and SPO-C consistently outperform the two-stage PTO framework for control ratios between 20\% and 70\%, achieving a maximum accuracy improvement of 9.22\% in the Uniform distribution scenario for the 100 $\times$ 100 network. However, this performance advantage reverses when the control ratio reaches 80\%, where SPO-A underperforms PTO in nearly all scenarios. We attribute this to precision errors in the alternating differentiation method, as the optimization module's influence becomes dominant at higher control ratios. 2) The SPO-A and SPO-C frameworks maintain comparable accuracy levels across the 20\%-70\% control ratio range, with observed deviations remaining minimal (peak deviation: 5.23\%). 3) Overall, the SPO-A framework demonstrates robust consistency in matching accuracy across all tested network sizes and distribution scenarios, replicating the reliability observed in Case A.

\begin{figure}[h]
    \centering
    \includegraphics[width=0.9\linewidth]{Figure/line_plot_cdc.png}
    \caption{The improvement in matching performance of SPO-A with SPO-C and PTO frameworks under Uniform, Gaussian, and Gaussian Mixture distributions in three large-scale networks, using the matching RMSE of the PTO framework as a reference.}
    \label{fig: lineplot cdc}
\end{figure}

Table \ref{tab: efficiency cdc} summarizes the computational efficiency across the three network sizes ($60\times60$, $80\times80$, and $100\times100$) and target distributions (Uniform, Gaussian, and Gaussian Mixture). One can observe minimal variation in total running time for SPO-A under different distributions within the same network size (0.91, 1.09, and 1.49 for $60\times60$, $80\times80$, and $100\times100$ networks, respectively). However, both SPO-A and SPO-C exhibit substantial running time expansion with the increment of network sizes, escalating from approximately 32,000 to 97,000. Notably, SPO-A maintains consistent superiority over SPO-C, demonstrating improvements of 2.2\%, 5.7\%, and 2.4\% in total running time for the respective network sizes. The findings align with the results in Case A. 

Overall, the consistent matching and efficiency performance of SPO-A across both Case A and Case B not only demonstrates the robustness and effectiveness of the SPO-A framework in terms of both matching accuracy and computational efficiency, but also establishes the generalization of the proposed SPO-A framework across different scales of problems formulated in quadratic programming.

\begin{table}[H]
    \centering
    \caption{Computational efficiency comparison of different target sensing distributions in three large-scale networks in Case B. (Efficiency is reported in total running time ($\times10^3$ sec).)} 
    \label{tab: efficiency cdc}
    \begin{tabular}{c|cccccc}
    \toprule
       \multirow{2}{*}{Distribution}& \multicolumn{2}{c}{$60\times60$}  & \multicolumn{2}{c}{$80\times80$} & \multicolumn{2}{c}{$100\times100$}  \\
        \cmidrule(lr){2-3} \cmidrule(lr){4-5} \cmidrule(lr){6-7}
        & \textbf{SPO-A} & SPO-C & \textbf{SPO-A} & SPO-C & \textbf{SPO-A} & SPO-C \\
        \midrule
        Uniform & 32.39 & 33.12 & 54.82 & 56.38 & 95.85 & 98.24	\\
        Gaussian &33.53& 33.98&	56.97 &	58.43 &	97.43 &	95.33 \\
        Gaussian Mixture  &34.18& 34.23 & 56.23 & 59.66 & 97.69 & 97.95 \\
        \midrule
        Mean  & \textbf{33.37} $\pm$ 0.91 & 33.77 $\pm$ 0.58 & \textbf{56.01} $\pm$1.09 & 58.15 $\pm$1.49 & \textbf{96.09} $\pm$1.49 & 97.17 $\pm$ 1.60\\  
        \bottomrule
    \end{tabular}
\end{table}

\section{Conclusions}
\label{sec: conclusion}

This paper presents an end-to-end SPO framework coupled with an alternating differentiation method (SPO-A) for vehicle relocation problems in mobile sensing, aimed at enhancing sensing efficiency with limited budgets. The proposed SPO-A framework integrates a constrained QP optimization layer in the neural network and derives an explicit alternating differentiation method based on the ADMM. The matrix-based formulation in optimization enables seamless integration of the optimization layer into deep learning frameworks, facilitating efficient training and deployment for vehicle relocation problems in VCS. The unrolling alternating differentiation approach within the optimization layer enables effective backpropagation for large-scale networks, essential for training deep learning models.

The effectiveness of the proposed SPO-A framework is validated through two real-world experiments conducted in both mid-size and large-scale networks in Hong Kong and Chengdu, China. Results demonstrate that the SPO-A framework surpasses most benchmarks set by the SPO with implicit differentiation techniques (SPO-C) and PTO in overall matching accuracy,  particularly in large-scale networks. Specifically, the SPO-A framework also outperforms the SPO-C in both matching accuracy and computational efficiency in most scenarios, highlighting the superior scalability of SPO-A for high-dimensional problems. The SPO-A framework outperforms PTO by providing more stable predictions during low demand variation and, crucially, by reducing overestimation in high-demand downtown areas during peak hours, which compensates for its peripheral underestimation and leads to superior overall matching performance. Finally, sensitivity analyses validate the robustness of the SPO-A framework against uncertainties in target distribution and varying control ratios, underscoring its applicability across relocation problems in diverse VCS scenarios.

Future directions of this research could focus on exploring further explicit differentiation methods specifically tailored for optimization layers, aimed at significantly enhancing computational efficiency, particularly in large-scale networks. Additionally, we aim to extend the application of the SPO framework to encompass other critical traffic management scenarios, such as vehicle routing, which presents an opportunity to tackle more intricate real-world challenges effectively. These advancements would not only expand the applicability of the SPO framework but also contribute to advancing the state-of-the-art in logistics and intelligent transportation systems.

\section*{Acknowledgments}
The work described in this paper is supported by grants from the Research Grants Council of the Hong Kong Special Administrative Region, China (Project No. PolyU/25209221 and PolyU/15206322) and a grant from the Otto Poon Charitable Foundation Smart Cities Research Institute (SCRI) at the Hong Kong Polytechnic University (Project No. P0043552). The contents of this article reflect the views of the authors, who are responsible for the facts and accuracy of the information presented herein. 

\clearpage
\appendix

\section{Notations}
\label{sec: appendix A}

Table \ref{apd: list of notations} presents the notations in the paper, and Table \ref{apd: variable vectorization} summarizes the dimensions of the vectorized variables for the SPO framework.
\setcounter{table}{0}
\renewcommand{\thetable}{A\arabic{table}} 

\begin{table}[H]
    \centering    
    \caption{List of notations.}
    \label{apd: list of notations}
    \resizebox{\textwidth}{!}{
    \begin{tabular}{p{2cm} p{14cm}}
        \toprule
        $I$ & The set of origin grids\\
        $J$ & The set of destination grids, $|I| = |J| = N$.\\
        $T$ & The set of all time intervals.\\
        $V$ & The set of vehicle class. \\
        \midrule
        \multicolumn{2}{c}{\textbf{Indices}}\\
        \midrule
        $v$ & The index of vehicle type, $v = \{a, c, f \}$. $a$ represents all vehicles, $c$ represents controllable DVs, $f$ represents non-dedicated free vehicles. \\ 
        $\tau$ & The index of the time interval. \\
        $k$ & The index of iterations. \\   
        $n$ & The index of constraints. \\
        \midrule
        \multicolumn{2}{c}{\textbf{Variables as scalars}}\\
        \midrule
        $D_{v,i}^{\tau}$ & The demand of vehicle type $v$ in grid $i$ at time $\tau$. \\
        $A_{f,i}^{\tau}$ & The adjacent information for NDVs in grid $i$ at time $\tau$. \\
        $H_{f,i}^{\tau}$ & The hidden information for NDVs in grid $i$ at time $\tau$. \\
        $x_{v,ij}^{\tau}$ & The vehicle flow for vehicle type v from origin $i$ to destination $j$ at time $\tau$.\\
        $w_{c,ij}^{\tau}$ & The incentive cost  from origin $i$ to destination $j$ for DVs at time $\tau$.\\
        $m_{c,ij}^{\tau}$ & The travel time from origin $i$ to destination $j$ for DVs at time $\tau$.\\
        \midrule
        \multicolumn{2}{c}{\textbf{Variables/Parameters as tensors}}\\
        \midrule
        $\mathbf{y}_k$ & The one-dimension vector of   vehicle flow in the $k^{th}$ iteration. \\
        $\mathbf{D}_v^\tau$ & The spatial distribution for vehicle type $v$ at time $\tau$. \\
        $\mathbf{T}_a^{\tau}$ & The target distribution of all vehicles $a$ at time $\tau$. \\
        $\mathbf{s}_{k}$ & The slack variables in $k^{th}$ iteration. \\
        $\mathbf{\mu}_{k}$ & The dual variables in $k^{th}$ iteration. \\  
        \midrule
        \multicolumn{2}{c}{\textbf{Parameters}} \\
        \midrule
        $\mathscr{L}$ & The loss function. $\mathscr{L}_1$ is the prediction loss, $\mathscr{L}_2$ is the matching loss, and $\mathscr{L}_{SPO}$ is the SPO loss. \\
        $w$ & The weight. $w_1$ is the weight of the prediction loss in the SPO function, and $w_2$ is the weight of the matching loss in the SPO framework. \\
        $\mathbf{w}_p$ & The weight in the prediction model. \\
        $\delta$ & The time length for each time interval. \\
        $\alpha$ & The Look back window in the prediction module. \\
        $\mathbf{C}$ & The incentive cost vector. \\
        $R$ & The total budget of the incentive cost. \\ 
        $K$ & The total number of alternating layers. \\ 
        $\xi$ & The convergence threshold. \\
        $\gamma$ & The control ratio. \\
        $\rho$ & The penalty term. \\
        \toprule
    \end{tabular}}
\end{table}

\begin{table}[H]
    \centering    
    \caption{Dimension of the vectors in the SPO framework.}
    \label{apd: variable vectorization}
    \begin{tabular}{cccccc}
        \toprule
          Vector & Dimension & Type & Vector & Dimension & Type \\
        \midrule
        \multicolumn{6}{c}{\textbf{Variables}}\\
         $\mathbf{y}$ & $\mathbb{R}^{N^2}$& Vector \\
         \multicolumn{6}{c}{\textbf{Parameters}}\\
         $\mathbf{A}$ & $\mathbb{R}^{N \times N^2}$ & Sparse Matrix & $\mathbf{B}$ & $\mathbb{R}^{N \times N^2}$ & Sparse Matrix\\
          $\mathbf{C}$ & $\mathbb{R}^{N^2}$ & Vector & 
         ${\mathbf{D}_v^{\tau}}$&$\mathbb{R}^{N}$& Vector \\
          $\mathbf{P}$ & $\mathbb{R}^{N^2 \times N^2}$ & Sparse Matrix &  
          $\mathbf{q}$ & $\mathbb{R}^{N^2}$ & Vector \\  
          $\mathbf{G_1}$ & $\mathbb{R}^{N \times N^2}$ & Sparse Matrix &          $\mathbf{s_1}$ & $\mathbb{R}^{N}$ & Vector \\    
          $\mathbf{G_2}$ & $\mathbb{R}^{N^2 \times N^2}$ & Sparse Matrix &          $\mathbf{s_2}$ & $\mathbb{R}^{N^2}$ & Vector \\  
          $\mathbf{G_3}$ & $\mathbb{R}^{1 \times N^2}$ & Vector &           $\mathbf{s_3}$ & $\mathbb{R}^{1}$ & Constant \\  
          $\mathbf{G_4}$ & $\mathbb{R}^{N^2 \times N^2}$ & Sparse Matrix&  
          $\mathbf{s_4}$ & $\mathbb{R}^{N^2}$ & Vector \\  
          $\mathbf{h_1}$ & $\mathbb{R}^{N}$ & Vector &           $\mathbf{\mu_1}$ & $\mathbb{R}^{N}$ & Vector \\  
          $\mathbf{h_2}$ & $\mathbb{R}^{N^2}$ & Vector &           $\mathbf{\mu_2}$ & $\mathbb{R}^{N^2}$ & Vector\\ 
          $\mathbf{h_3}$ & $\mathbb{R}^{1}$ & Vector &           $\mathbf{\mu_3}$ & $\mathbb{R}^{1}$ & Constant \\
          $\mathbf{h_4}$ & $\mathbb{R}^{N^2}$ & Vector & 
          $\mathbf{\mu_4}$ & $\mathbb{R}^{N^2}$ & Vector \\
        \toprule
    \end{tabular}
\end{table}

\section{Hyper-parameters in different prediction modules.}
\label{sec: appendix B}
Table \ref{apd: hyperparameter} lists the hyper-parameters used in various prediction models.

\setcounter{table}{0}
\renewcommand{\thetable}{B\arabic{table}} 

\begin{table}[H]
    \centering
    \caption{Hyper-parameters of different prediction modules.}
    \label{apd: hyperparameter}
    \addtolength{\tabcolsep}{12pt}
    \begin{tabular}{ccc}
    \toprule
       Model  & Hyper-parameter & Value \\
    \midrule
        \multirow{4}{*}{TGCN} & Number of hidden layers & 2\\
        & Number of channels in each layer & 64, 32 \\
        & Kernel size & 2 \\
        & Dropout & 0.2 \\
        & Activation function & ReLU \\
    \midrule
        \multirow{3}{*}{DCRNN} & Number of hidden layers & 2\\
        & Number of channels in each layer & 64, 32 \\
        & Dropout & 0.2 \\
        & Activation function & ReLU \\
    \midrule
        \multirow{5}{*}{ASTGCN} & Number of hidden layers & 2\\
        & Number of channels in each layer & 32, 16 \\
        & Number of heads & 2 \\
        & Dropout & 0.2 \\
        & Activation function & ReLU \\
        \bottomrule
    \end{tabular}
\end{table}

\newpage
\bibliography{main_clean}

\end{document}